%% file: main.tex
\documentclass[sigconf]{acmart}

\AtBeginDocument{%
  }

\setcopyright{acmlicensed}
\copyrightyear{2018}
\acmYear{2018}
\acmDOI{XXXXXXX.XXXXXXX}
\acmConference[Conference acronym 'XX]{Make sure to enter the correct
  conference title from your rights confirmation email}{June 03--05,
  2018}{Woodstock, NY}
\acmISBN{978-1-4503-XXXX-X/2018/06}

\usepackage[utf8]{inputenc} 
\usepackage[T1]{fontenc}    
\usepackage{hyperref}       
\usepackage{url}            
\usepackage{booktabs}       
\usepackage{amsfonts}       
\usepackage{nicefrac}       
\usepackage{microtype}      
\usepackage{xcolor}         
\usepackage{makecell}
\usepackage{amsthm} 
\usepackage{multirow}

\usepackage{amsmath}
\usepackage{mathtools}
\usepackage{amsthm}
\usepackage{diagbox}
\usepackage{tabularx}

\usepackage{booktabs}
\usepackage{subcaption}
\usepackage{bm}
\usepackage{graphicx}
\usepackage{tikz}
\usetikzlibrary{positioning,arrows.meta,calc,shapes.misc,fit}
\usepackage{multirow}
\usepackage{colortbl}
\usepackage{rotating}
\usepackage{makecell}
\usepackage{wrapfig}
\usepackage{enumitem}
\usepackage{pifont}
\usepackage{makecell}
\usepackage[ruled,vlined,linesnumbered]{algorithm2e}

\usepackage[most]{tcolorbox}

\usepackage{multicol}
\usepackage{setspace}

\usepackage{titletoc}

\definecolor{best}{rgb}{1,  0,  0}
\definecolor{second}{rgb}{0,  0,  0.859}
\definecolor{yellowhl}{rgb}{1, 0.97, 0.7}

\usepackage{hyperref}
\SetCommentSty{footnotesize}
\SetKwComment{Comment}{$\triangleright$ }{}
\SetSideCommentRight

\theoremstyle{remark}

\def\method{\texttt{TimeFKA}~}

\newcommand{\pctb}[1]{{\scriptsize\textcolor{blue}{($\downarrow$#1)}}}
   \newcommand{\pctr}[1]{{\scriptsize\textcolor{red}{($\uparrow$#1)}}}  

\definecolor{BlueGreen}{RGB}{105,204,207}
\definecolor{RedOrange}{RGB}{255,69,0}

\definecolor{darkblue}{rgb}{0, 0, 0.5}
\hypersetup{colorlinks=true, citecolor=darkblue, linkcolor=darkblue, urlcolor=darkblue}

\definecolor{HeaderBg}{HTML}{E9E9F7}
\definecolor{RowBg}{HTML}{F4F5FC}
\definecolor{RuleGray}{HTML}{555555}

\definecolor{ExactColor}{HTML}{F3B36B}     
\definecolor{MultiColor}{HTML}{8E9AC8}     
\definecolor{SeqColor}{HTML}{80B9B2}       
\definecolor{PrivColor}{HTML}{b25d25}      
\definecolor{RubricColor}{HTML}{C47AA5}    

\definecolor{myred}{RGB}{184,26,15}
\definecolor{mydarkred}{rgb}{0.6,0,0}
\definecolor{frontierblue}{HTML}{EAF3FF}
\definecolor{frontierblueavg}{HTML}{F4F9FF}

\definecolor{bestcolor}{HTML}{800080}
\definecolor{secondcolor}{HTML}{2da02d}

\newcommand{\firstres}[1]{{\textcolor{bestcolor}{\textbf{#1}}}}
\newcommand{\secondres}[1]{{\textcolor{secondcolor}{\underline{#1}}}}

\tcolorboxenvironment{proposition}{
  notitle,
  rounded corners,
  colframe=darkgray,
  colback=blue!5,
  boxrule=0.75pt,
  boxsep=0pt,
  left=0.2cm,
  right=0.2cm,
  enhanced,
  shadow={1pt}{-1pt}{0pt}{opacity=0.5,gray},
  toprule=0.75pt,
  before skip=0.65em,
  after skip=0.75em
}
\tcolorboxenvironment{lemma}{
  notitle,
  rounded corners,
  colframe=darkgray,
  colback=blue!5,
  boxrule=0.75pt,
  boxsep=0pt,
  left=0.2cm,
  right=0.2cm,
  enhanced,
  shadow={1pt}{-1pt}{0pt}{opacity=0.5,gray},
  toprule=0.75pt,
  before skip=0.65em,
  after skip=0.75em
}
\tcolorboxenvironment{corollary}{
  enhanced, breakable,
  colback=gray!6, colframe=gray!55,
  boxrule=0.4pt, arc=1.5pt,
  left=4pt, right=4pt, top=3pt, bottom=3pt,
  before skip=6pt, after skip=6pt,
}
\tcolorboxenvironment{remark}{
  enhanced, breakable,
  colback=gray!6, colframe=gray!55,
  boxrule=0.4pt, arc=1.5pt,
  left=4pt, right=4pt, top=3pt, bottom=3pt,
  before skip=6pt, after skip=6pt,
}

\renewcommand\footnotetextcopyrightpermission[1]{}

\begin{document}

\title{Decouple to Reuse: Fine-Grained Knowledge Augmentation for Time Series Foundation Models}

\author{Yuqian Wu $^*$}
\affiliation{%
  \institution{DSA Thrust, The Hong Kong University of Science and Technology (Guangzhou)}
  \institution{Solar System of OVB, Tencent}
  \city{Guangzhou}
  \country{China}
}
\email{ywu188@connect.hkust-gz.edu.cn}

\author{Qingxiang Liu $^*, ^{\dag}$}
\thanks{$^*$ Equal contribution.}
\affiliation{%
  \institution{INTR Thrust, The Hong Kong University of Science and Technology (Guangzhou)}
  \city{Guangzhou}
  \country{China}
}
\email{qingxiangliu737@gmail.com}

\author{Yuxin Guo}
\affiliation{%
    \institution{Beijing Information Science and Technology University}
    \city{Beijing}
  \country{China}
}
\email{2025040181@bistu.edu.cn}

\author{Weifeng Su}
\affiliation{%
    \institution{Beijing Normal-Hong Kong Baptist University}
    \city{Zhuhai}
  \country{China}
}
\email{wfsu@bnbu.edu.cn}

\author{Jian Li}
\affiliation{%
    \institution{Solar System of OVB, Tencent}
    \city{Shenzhen}
  \country{China}
}
\email{loucasli@tencent.com}

\author{Pengfei Xu}
\affiliation{%
    \institution{Solar System of OVB, Tencent}
    \city{Shenzhen}
  \country{China}
}
\email{luciferxu@tencent.com}

\author{Xi Chen}
\affiliation{%
    \institution{Solar System of OVB, Tencent}
    \city{Shenzhen}
  \country{China}
}
\email{jasonxchen@tencent.com}

\author{Yutao Yue}
\affiliation{%
    \institution{INTR Thrust, The Hong Kong University of Science and Technology (Guangzhou)}
    \city{Guangzhou}
  \country{China}
}
\email{yutaoyue@hkust-gz.edu.cn}

\author{Yuxuan Liang$^{\dag}$}
\thanks{$^{\dag}$ Corresponding authors.}
\affiliation{%
  \institution{INTR \& DSA Thrust, The Hong Kong University of Science and Technology (Guangzhou)}
  \city{Guangzhou}
  \country{China}
}
\email{yuxuanliang@hkust-gz.edu.cn}



\received{20 February 2007}
\received[revised]{12 March 2009}
\received[accepted]{5 June 2009}

\input{section/00_abstract}

\maketitle
\input{section/01_introduction}

\input{section/05_related_work}
\input{ section/03_method}

\input{ section/04_experiment}
\input{ section/06_conclusion}


\section*{Ethical Considerations \& AI Disclosure}
This study uses publicly available benchmark datasets for model training and evaluation and does not involve private user data or personal information.
The proposed framework is intended to support research on time series forecasting and web mining.
We acknowledge the use of Large Language Models during the preparation of this manuscript for the purpose of language polishing and grammar correction. All core scientific contributions are developed solely by the authors.
\balance

\bibliographystyle{ACM-Reference-Format}
\bibliography{reference}




\appendix
\input{appendix/0_appendix_list}

\end{document}

%% file: section/00_abstract.tex
\begin{abstract}

Time series forecasting is critical for demand prediction and resource management in modern Web services, yet applying pretrained Time Series Foundation Models (TSFMs) to downstream tasks remains challenging due to domain shifts.
Existing retrieval-augmented approaches typically retrieve coarse-grained window contexts and may introduce context-specific noise.
To address this problem, we argue that temporal patterns should be reused at a decomposition level, as effective trend and seasonal dynamics may originate from different contexts.
We propose \method, a two-stage \underline{F}ine-grained \underline{K}nowledge \underline{A}ugmentation framework that decouples and consolidates pre-training knowledge into compact, task-shared residual codebooks to enhance frozen TSFMs. 
Specifically, Stage I constructs dual-branch residual codebooks for decoupled trend and seasonal signals, while Stage II dynamically retrieves and merges these fine-grained patterns into target representations via task-adaptive aggregation and gate fusion.
Extensive experiments across eight real-world benchmarks with four representative TSFM backbones demonstrate the superior effectiveness and computational efficiency of \method, while further validating the strong transferability of the residual codebooks across different backbones.
The source code is available at the anonymous repository: {\url{https://github.com/yuppielqx/TSFM-Component-Token-Adapter}}.
\end{abstract}

%% file: section/01_introduction.tex
\section{Introduction}
\label{sec:intro}

The operations of modern Web services underline the important role of time series forecasting, from demand evaluation on online platforms to cloud resource provisioning and intelligent transportation management~\cite{ma2026quizsf,jiang2026cross,fan2026automated,zhang2026semixer,deng2025lm, deng2026codeblock,wu2026lifeside,wu2026back}.
The diversity of these services motivates the development of unified forecasting models that can be reused across various applications.
Recent researches have shifted from training task-specific models from scratch~\cite{zhou2021informer,wu2021autoformer,wu2022timesnet,wu2025beyond} to developing Time Series Foundation Models (TSFMs)~\cite{das2024decoder,ansari2024chronos,ekambaram2024tiny,ansari2025chronos,liu2026sundial,liu2026rationale,chen2026urbanfm}, which learn transferable temporal representations through pre-training on diverse temporal corpora.
However, service-specific demand levels, usage cycles, and workload dynamics can differ from the patterns captured during pre-training, limiting the zero-shot forecasting performance of TSFMs in Web deployments~\cite{he2023domain,zhao2025proactive}.

Existing efforts in enhancing the downstream forecasting performance of TSFMs broadly follow two complementary paradigms.
\begin{itemize}[leftmargin=*,itemsep=0.5em,parsep=1pt]
    \item[\ddag] \textbf{Parametric Adaptation}:
    These approaches tailor pre-trained TSFMs to downstream tasks through full or parameter-efficient fine-tuning~\cite{gao2024units,ruan2025st,qiao2026multi}.
    For example, UniTS~\cite{gao2024units} learns task-specific prompt tokens while keeping the pre-trained backbone frozen.
    Although effective, adapting to multiple deployment domains often requires separate optimization and maintenance of domain-specific parameters, increasing the overall adaptation burden.

    \item[\ddag] \textbf{Knowledge Augmentation}:
    These approaches enrich forecasting with temporal knowledge from auxiliary data sources~\cite{tire2024retrieval,ning2026ts,zhou2026stationarity,chen2025select,qin2025cora}.
    For instance, TS-RAG~\cite{ning2026ts} retrieves historical contexts from a knowledge base constructed from pre-training data and incorporates their associated future segments into the TSFM representation through an adaptive retrieval mixer.
    However, auxiliary information may contain both reusable temporal regularities and context-specific variations. Incorporating them together can introduce noisy patterns that do not match the current input.
\end{itemize}

\begin{figure*}[!t]
  \centering
  \includegraphics[width=\textwidth]{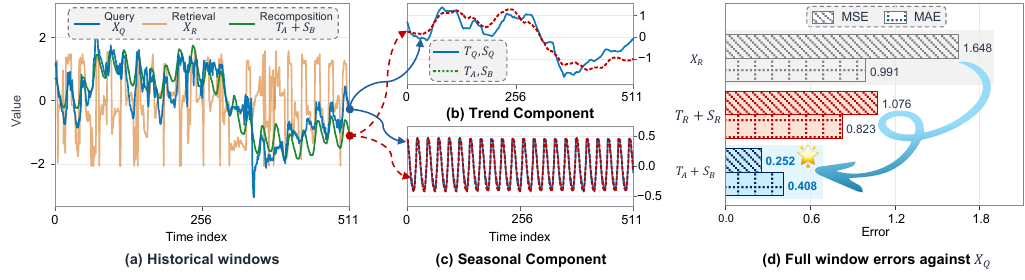}
    \caption{
Motivating cases for fine-grained knowledge augmentation. 
(a) Comparison of the query ($X_Q$), a retrieved window ($X_R$) in TS-RAG, and a recomposed reference ($T_A + S_B$). 
(b--c) Comparison of the query components ($T_Q$ and $S_Q$) with the matched trend ($T_A$) and seasonal ($S_B$) patterns selected from different pre-training contexts.
(d) Window-level MSE and MAE measured against $X_Q$, where $T_R$ and $S_R$ are decoupled from $X_R$.
Combining trend and seasonal patterns from different contexts yields a closer match to the query than directly retrieving an entire window.
}
\label{fig:paradigm}
\end{figure*}

We focus on knowledge augmentation and investigate how temporal knowledge from pre-training corpora can be explicitly organized and reused to enhance a frozen TSFM.
Our key observation is that \textbf{useful temporal patterns for a downstream \textit{query} may be distributed across different contexts} selected from the pre-training dataset.
As illustrated in Fig.~\ref{fig:paradigm}(a), the distribution of a window retrieved by TS-RAG can differ substantially from that of the query.
Nevertheless, the trend pattern from one selected context and the seasonal pattern from another can separately align with the corresponding query components, as shown in Fig.~\ref{fig:paradigm}(b) and (c).
Combining these patterns produces a closer match to the observed query context, as reflected by the lower window-level matching errors in Fig.~\ref{fig:paradigm}(d).
These cases suggest that whole-window similarity can obscure the reuse potential of constituent temporal patterns.
This motivates reorganizing pre-training knowledge around recurring patterns rather than their source windows.
We therefore seek to learn a compact, shared bank of trend and seasonal patterns, enabling task-specific combinations of these patterns to condition a frozen TSFM for downstream forecasting.

To this end, we propose \method, a two-stage Fine-grained Knowledge Augmentation framework that consolidates temporal knowledge from pre-training corpora into shared pattern banks to enhance frozen TSFMs.
In Stage I, \method decomposes pre-training contexts into trend and seasonal components and encodes them using a frozen TSFM backbone.
We then learn shared trend and seasonal codebooks for all downstream tasks by residual reconstruction to consolidate the reusable knowledge into compact pattern banks.
In Stage II, we retrieve the trend and seasonal patterns for each downstream input from the cultivated codebooks.
We combine these reusable patterns with identity embeddings through task-adaptive aggregation and then augment the raw temporal representation through gate fusion.
\method complements target-specific parameter adaptation with explicit access to shared pattern banks learned from pre-training data.
Compared with window-based retrieval, \method enables fine-grained knowledge reuse by respectively retrieving trend and seasonal patterns to augment the prediction performance of pretrained TSFMs.
During downstream adaptation and inference, pattern retrieval operates on the compact codebooks without accessing the original pre-training windows.

The main contributions of this work are summarized as follows.
\begin{itemize}[leftmargin=*]
    \item \textbf{Fine-grained Knowledge Augmentation.}
    We highlight that useful trend and seasonal patterns for a downstream query may occur in different pre-training contexts. This observation motivates organizing temporal knowledge as independently accessible patterns that can be reused beyond their original window-level retrieval.

    \item \textbf{Two-stage Knowledge Augmentation Framework.}
    We propose \method, which firstly learns shared residual codebooks to separately consolidate trend and seasonal patterns for all downstream tasks and then improves the prediction performance of frozen TSFMs by task-specific knowledge augmentation. 
    
    \item \textbf{Extensive Empirical Validation.}
    We conduct performance evaluation on eight real-world forecasting datasets with four pretrained TSFMs. 
    The empirical results validate the effectiveness, efficiency, and scalability of our proposed \method.
\end{itemize}

%% file: section/05_related_work.tex
\section{Related Work}
In this section, we review the efforts on pretrained TSFMs and the following enhancement approaches, including parametric adaptation and knowledge augmentation.

\noindent\textbf{Time Series Foundation Models.}
TSFMs learn transferable temporal representations through large-scale pre-training to support forecasting across datasets and domains~\cite{garza2024timegpt1,rasul2024lagllama}.
Existing models adopt several backbone architectures.
\textit{Encoder-only} models, including MOMENT~\cite{goswami2024moment}, the original Moirai~\cite{woo2024unified}, and Chronos-2~\cite{ansari2025chronos}, encode temporal contexts into representations used by downstream prediction heads.
\textit{Decoder-only} models, such as TimesFM~\cite{das2024decoder}, Timer and Timer-XL~\cite{liu2024timer,liu2025timerx}, and Sundial~\cite{liu2026sundial}, use causal attention to model temporal dependencies for generative forecasting.
\textit{Encoder--decoder} models, such as Chronos and Chronos-Bolt~\cite{ansari2024chronos}, separate historical context encoding from future decoding.
Beyond Transformer architectures, \textit{MLP-based} models such as TTM~\cite{ekambaram2024tiny} employ lightweight mixing blocks for efficient forecasting.
However, these TSFMs' zero-shot performance can remain limited when downstream temporal dynamics differ from those encountered during pre-training~\cite{ye2026empowering}, motivating adaptation and knowledge augmentation approaches.


\noindent\textbf{Parametric Adaptation.}
Parametric adaptation tailors pre-trained TSFMs to downstream tasks through full fine-tuning or parameter-efficient updates~\cite{gao2024units,ruan2025st,qiao2026multi}.
For example, UniTS~\cite{gao2024units} learns task-specific prompts for a frozen backbone, while ST-LoRA~\cite{ruan2025st} introduces low-rank adaptation for spatio-temporal forecasting.
These approaches provide task-specific flexibility, although deployment across multiple domains may require separate optimization and maintenance of adapted parameters.

\noindent\textbf{Knowledge Augmentation.}
Knowledge augmentation enriches forecasting with auxiliary information, including retrieved temporal references~\cite{tire2024retrieval,ning2026ts,zhou2026stationarity}, exogenous covariates~\cite{chen2025select,qin2025cora}, and parameterized memories~\cite{lyu2026ts,yu2026memts}.
TS-RAG~\cite{ning2026ts} retrieves historical contexts and their associated futures and integrates them with TSFM representations through an adaptive retrieval mixer.
TS-Memory~\cite{lyu2026ts} distills retrieval-induced forecasting corrections into a lightweight memory adapter, whereas MEMTS~\cite{yu2026memts} internalizes domain-specific temporal dynamics in learnable latent prototypes.
These approaches differ in how knowledge is represented and accessed.
\method organizes pre-training knowledge into separate trend and seasonal residual codebooks and learns their task-adaptive aggregation and fusion with a frozen TSFM.
This design enables fine-grained knowledge reuse while preserving the shared banks during downstream adaptation.

%% file: section/03_method.tex
\begin{figure*}[!t]
    \centering
    \includegraphics[width=\linewidth]{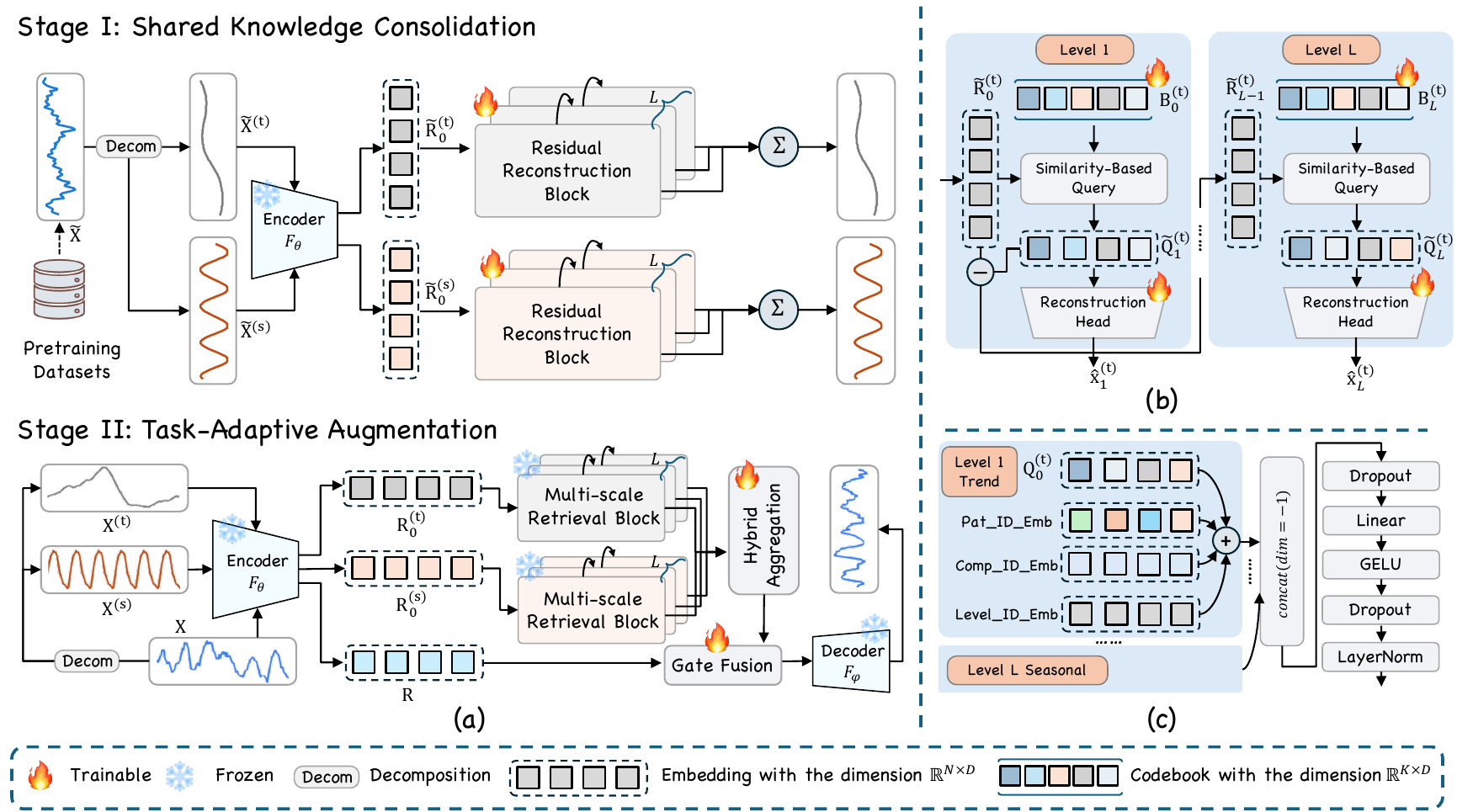}
    \caption{Overall framework of \method. (a) Two stages: Shared Knowledge Consolidation and Task-Adaptive Augmentation.
    (b) Residual reconstruction block in the trend branch.
    (c) Hybrid Aggregation.}
    \label{method}
\end{figure*}

\section{Methodology}

Let $\mathbf{x}_t \in \mathbb{R}^d$ denote the observations from $d$ sensors in a web-service system at the time stamp $t$.
In the literature of time series forecasting, we use the preceding $T$ observations as the historical input $\mathbf{X}_t =
[\mathbf{x}_{t-T+1},\ldots,\mathbf{x}_t]^\top \in\mathbb{R}^{T\times d}$ and predict the following $H$ observations $\mathbf{Y}_t = [\mathbf{x}_{t+1},\ldots,\mathbf{x}_{t+H}]^\top \in\mathbb{R}^{H\times d}$, where $T$ and $H$ denote the context and forecast lengths, respectively.
We adopt the channel-independent strategy to each paired $(\mathbf{X}_t, \mathbf{Y}_t)$~\cite{nie2022patchtst}.
For simplicity, we consider a single channel and omit the subscript index to obtain $\mathbf{X} \in \mathbb{R}^{T\times 1}$ and $\mathbf{Y}\in \mathbb{R}^{H\times 1}$.
The historical observation $\mathbf{X}$ can be decomposed into \textit{trend} $\mathbf{X}^{({\rm t})} \in \mathbb{R}^{T \times 1}$ and \textit{seasonal} components $\mathbf{X}^{({\rm s})} \in \mathbb{R}^{T\times 1}$ \cite{wu2021autoformer,wu2022timesnet}.



The overall framework of \method is shown in Fig.~\ref{method}.
Specifically, \method includes two stages: Stage I aims to consolidate reusable knowledge from the pretrained corpus into residual codebooks shared across all downstream forecasting tasks;
Stage II enhances the prediction performance of TSFMs by task-adaptive augmentation.

\subsection{Stage I: Shared Knowledge Consolidation}
\label{sec:codebook_learning}
In the knowledge augmentation-based methods, diverse formats of temporal knowledge are organized from the pre-training corpus, such as time series samples in TS-RAG~\cite{ning2026ts} and prediction quantiles~\cite{lyu2026ts}.
However, the coarse-grained knowledge may introduce context-specific variations and perturb downstream forecasting.
In this stage, we aim to cultivate the underlying temporal patterns from the trend and seasonal perspectives, as the basic retrieval units.
Specifically, we decompose sampled pre-training contexts into trend and seasonal components and learn separate residual quantization codebooks, which serve as fine-grained retrieval units for downstream knowledge augmentation.

Following TS-RAG~\cite{ning2026ts}, we first uniformly sample time-series instances from the pretraining corpus of Chronos~\cite{ansari2024chronos} and only retain their input contexts to construct $\mathcal{D}_{\rm pre}$, but exclude the future segments.
Each segment $\widetilde{\mathbf{X}} \in \mathcal{D}_{\rm pre}$ is preprocessed by the channel-independent strategy, thus $\widetilde{\mathbf{X}} \in \mathbb{R}^{T\times 1}$.
We also decompose $\widetilde{\mathbf{X}}$ into the trend $\widetilde{\mathbf{X}}^{(\rm t)} \in \mathbb{R}^{T\times 1}$ and seasonal components $\widetilde{\mathbf{X}}^{(\rm s)} \in \mathbb{R}^{T\times 1}$, which work as the reconstruction targets to cultivate temporal codebooks.

For each component ${\rm m} \in \{ {\rm s}: seasonal, {\rm t}: trend\}$, the overall forward process can be formulated as
\begin{equation}
\begin{aligned}
\widetilde{\mathbf{R}}^{({\rm m})}_0
&=\texttt{sg}\!\left(F_\theta\!\left(\widetilde{\mathbf{X}}^{(\rm m)}\right)\right),\\
\bigl(\widetilde{\mathbf{R}}_l^{(\rm m)},\widehat{\mathbf{x}}_l^{(\rm m)}\bigr)
&=\texttt{RRB}_l^{(\rm m)}\bigl(\widetilde{\mathbf{R}}_{l-1}^{(\rm m)}\bigr),1\le l \le L , \\
\widehat{\mathbf{X}}^{(\rm m)}
&=\sum_{l=1}^{L}\widehat{\mathbf{x}}_l^{(\rm m)},
\end{aligned}
\label{eq:stage1_overview}
\end{equation}
where $\texttt{sg}(\cdot)$ represents the stop-gradient operation. $F_\theta$ denotes the frozen TSFM backbone, which divides $\widetilde{\mathbf{X}}^{(\rm m)}$ into non-overlapping patches with the length of $L_p$ and accordingly we can have $N=\lceil \frac{T-L_p}{L_p} \rceil +1$ patches.
We can have $\widetilde{\mathbf{R}}_0^{(\mathrm m)}\in\mathbb{R}^{N\times D}$ from the frozen TSFM backbone.
The \textit{residual reconstruction block} $\texttt{RRB}_l^{(\mathrm m)}$ returns the residual embedding for the next block and a contribution $\widehat{\mathbf{x}}_l^{(\mathrm m)}\in\mathbb{R}^{T\times 1}$ for reconstruction.

\noindent \textbf{Residual Reconstruction Block.}
Each $\texttt{RRB}_l^{(\mathrm m)}$ includes a trainable codebook $\mathbf{B}_l^{(\mathrm m)} \in\mathbb{R}^{K\times D}$ and a reconstruction head $G_l^{(\mathrm m)}$.
We quantize the residual embedding through row-wise similarity-based query:
\begin{equation}
    \widetilde{\boldsymbol I}_l^{(\rm m)} = \operatorname{NN}_{\mathrm{row}}
\left(
\widetilde{\mathbf R}_{l-1}^{(\rm m)},
\mathbf B_l^{(\rm m)}
\right),
\quad
\widetilde{\mathbf Q}_l^{(\rm m)}
=
\mathbf B_l^{(\rm m)}[\widetilde{\boldsymbol I}_l^{(\rm m)},:]
\in\mathbb R^{N\times D},
\label{eq_codebook_query}
\end{equation}
where $\operatorname{NN}_{\mathrm{row}}$ independently selects the most similar code vectors for each row under squared Euclidean distance and returns an index vector \(\boldsymbol I_l^{(\rm m)}\in\{1,\ldots,K\}^{N}\). The indexing operation obtains the selected code vectors correspondingly.

The quantized embedding $\widetilde{\mathbf Q}_l^{(\rm m)}$ is used for both reconstruction and residual refinement. 
For reconstruction, the level-specific head maps $\widetilde{\mathbf Q}_l^{(\rm m)}$ to the reconstruction contribution $\widehat{\mathbf x}_l^{(\rm m)}$:
\begin{equation}   
\begin{aligned}
G_l^{(\rm m)}\!\left(\widetilde{\mathbf Q}_l^{(\rm m)}\right)
&=
\widetilde{\mathbf Q}_l^{(\rm m)}\mathbf W_{\mathrm{rec},l}^{(\rm m)}
+
\mathbf 1_N
\left(\mathbf b_{\mathrm{rec},l}^{(\rm m)}\right)^{\top}
\in\mathbb R^{N\times L_p},
\\
\widehat{\mathbf x}_l^{(\rm m)}
&=
\texttt{Flatten}
\left(
G_l^{(\rm m)}\!\left(\widetilde{\mathbf Q}_l^{(\rm m)}\right)
\right),
\end{aligned}
\end{equation}
where \(\mathbf W_{\mathrm{rec},l}^{(\rm m)}\in\mathbb R^{D\times L_p}\), \(\mathbf b_{\mathrm{rec},l}^{(\rm m)}\in\mathbb R^{L_p}\), and \(\mathbf 1_N\in\mathbb R^{N\times1}\). 
The operator \(\texttt{Flatten}\) assembles the reconstructed patches in temporal order, following the boundary handling of the input patching procedure.
The residual embedding passed to the next block is updated as
\begin{equation}
    \widetilde{\mathbf R}_l^{(\rm m)} = \widetilde{\mathbf R}_{l-1}^{(\rm m)} - \texttt{sg}\!\left(\widetilde{\mathbf Q}_l^{(\rm m)}\right),
    \label{eq_residual_emb}
\end{equation}
The stop-gradient operation $\texttt{sg}$ applies only to the residual-update branch and the reconstruction branch remains differentiable with respect to the selected code vectors and reconstruction-head parameters. 
After \(L\) blocks, the level-wise reconstruction contributions are summed as in Eq. (1) to obtain \(\widehat{\mathbf X}^{(\rm m)}\). 

\noindent\textbf{Supervision Task.}
The supervision loss $\mathcal{L}_{\rm pre}^{(\rm m)}$ is formulated as
\begin{equation}
\begin{aligned}
\mathcal{L}_{\mathrm{align}}^{(\mathrm{m})}
&=
\frac{1}{L}
\sum_{l=1}^{L}
\left\|
\operatorname{sg}\left(
\widetilde{\mathbf{R}}_{l-1}^{(\mathrm{m})}
\right)
-
\widetilde{\mathbf{Q}}_l^{(\mathrm{m})}
\right\|_{2}^{2},
\\
\mathcal{L}_{\mathrm{rec}}^{(\mathrm{m})}
&=
\frac{1}{T}
\left\|
\widehat{\mathbf{X}}^{(\mathrm{m})}
-
\widetilde{\mathbf{X}}^{(\mathrm{m})}
\right\|_{2}^{2},
\\
\mathcal{L}_{\mathrm{pre}}^{(\mathrm{m})}
&=
\mathcal{L}_{\mathrm{rec}}^{(\mathrm{m})}
+
\lambda_{\mathrm{align}}
\mathcal{L}_{\mathrm{align}}^{(\mathrm{m})}.
\end{aligned}
\label{eq:stage1_losses}
\end{equation}
The alignment term encourages the selected code vectors in $\widetilde{\mathbf{Q}}_l^{(\mathrm{m})}$ to approximate the corresponding residual embeddings in $\widetilde{\mathbf{R}}_{l-1}^{(\mathrm{m})}$, which are treated as fixed targets by the stop-gradient operator.
The reconstruction term compares the combined reconstruction with the targets, jointly supervising all level-specific reconstruction heads.
The coefficient $\lambda_{\mathrm{align}}$ balances codebook alignment against reconstruction.
We minimize $\mathcal{L}_{\mathrm{pre}}^{(\mathrm{m})}$ over the trend or seasonal codebooks and their reconstruction heads while keeping the TSFM backbone frozen.
In this stage, we train the trend and seasonal residual codebooks in parallel. 

\subsection{Stage II: Task-Adaptive Augmentation}
\label{sec:component_tokens}

The residual codebooks learned in Stage I provide fine-grained temporal patterns, but similarity-based matching alone does not specify how the retrieved information should contribute to downstream forecasting.
Stage II therefore aims to adapt the use of the shared knowledge to the forecasting task.
Specifically, we aggregate trend and seasonal patterns across residual levels and integrate the resulting condition into the original TSFM representation through gated residual fusion.
This adapter is optimized under downstream forecasting supervision, while the codebooks and TSFM remain frozen.


Given a downstream input context $\mathbf{X}$, we obtain its trend and seasonal decompositions, denoted as $\mathbf{X}^{(\mathrm t)}$ and $\mathbf{X}^{(\mathrm s)}$ respectively.
The overall forward process in Stage II can be formulated as
\begin{equation}
\begin{aligned}
\mathbf{R}
=\texttt{sg}\!\left(F_\theta(\mathbf{X})\right) &,
\mathbf{R}_0^{(\mathrm m)} 
 =\texttt{sg}\!\left(F_\theta(\mathbf{X}^{(\mathrm m)})\right),
\\
\bigl(\mathbf{R}_l^{(\mathrm m)},\mathbf{Q}_l^{(\mathrm m)}\bigr)
&=\texttt{MRB}_l^{(\mathrm m)}\bigl(\mathbf{R}_{l-1}^{(\mathrm m)}\bigr),\\
\widehat{\mathbf{Y}}
=F_\varphi\bigl(\texttt{GateFuse}\bigl(\mathbf{R}&,\texttt{HyperAgg}\bigl(
\left\{\mathbf{Q}_l^{(\mathrm t)},\mathbf{Q}_l^{(\mathrm s)}\right\}_{l=1}^{L}
\bigr)\bigr)\bigr),\\
\end{aligned}
\label{eq:stage2_overview}
\end{equation}
where ${\rm m} \in \{ {\rm s}: seasonal, {\rm t}: trend\}$.
$\mathbf{R}$, $\mathbf{R}_0^{(\mathrm t)}$, and $\mathbf{R}_0^{(\mathrm s)}$ denote the frozen embeddings from the raw input, the trend and seasonal decompositions respectively.
The \textit{Multi-scale Retrieval Block} $\texttt{MRB}_l^{(\mathrm m)}$ queries the token vectors $\mathbf{Q}_l^{(\mathrm m)}$ and obtains the residual embedding $\mathbf{R}_l^{(\mathrm m)}$ for the next level.
We then perform a hybrid aggregation of these queried tokens and learnable identity embeddings. 
After gate fusion, the frozen decoder $F_\varphi$ of the adopted TSFM yields the prediction result $\widehat{\mathbf{Y}}$.

\noindent\textbf{Multi-scale Retrieval Block.}
For each component $\rm m \in \{\mathrm{s}, \mathrm{t}\}$, the $l$-th block obtains $\mathbf{Q}_l^{(\rm m)}$ and $\mathbf{R}_l^{(\rm m)}$ from $\mathbf{R}_{l-1}^{(\rm m)}$ following the retrieval and residual-update process in Eqs.~\eqref{eq_codebook_query} and~\eqref{eq_residual_emb} respectively, with the corresponding codebook $\mathbf{B}_l^{(\rm m)}$ kept frozen.
The updated residual $\mathbf{R}_l^{(\rm m)}$ is passed to the next level, while the retrieved representations $\left\{\mathbf{Q}_l^{(\mathrm{t})}, \mathbf{Q}_l^{(\mathrm{s})}\right\}_{l=1}^{L}$ are generated for hybrid aggregation.

\input{table/main_backbone}

\noindent\textbf{Hybrid Aggregation and Gate Fusion.}
The hybrid aggregation $\texttt{HyperAgg}$ operates on $\mathbf{Q}_l^{(\mathrm t)}$ and $\mathbf{Q}_l^{(\mathrm s)}$.
We first augment each matrix with trainable identity embeddings for the two components, $L$ quantization levels, and $N$ patch positions.
For the trend component:
\begin{equation}
\overline{\mathbf{Q}}_l^{(\mathrm t)} ={}\mathbf{Q}_l^{(\mathrm t)}+\mathbf{E}_{\mathrm{pos}} +\mathbf{1}_N \bigl(\mathbf{e}_{\mathrm{comp}}^{(\mathrm t)} \bigr)^\top
+\mathbf{1}_N \bigl( \mathbf{e}_{\mathrm{level}}^{(l)}\bigr)^\top,
\label{eq:identity_embeddings}
\end{equation}
where $\mathbf{e}_{\mathrm{comp}}^{(\mathrm t)},\mathbf{e}_{\mathrm{level}}^{(l)}\in\mathbb{R}^{D\times 1}$ are trainable component- and level-specific vectors respectively.
$\mathbf{E}_{\mathrm{pos}}\in\mathbb{R}^{N\times D}$ denotes the trainable patch-position embeddings.
We obtain $\overline{\mathbf{Q}}_l^{(\mathrm s)} \in \mathbb{R}^{N\times D}$ in the same way, replacing the trend component embedding $\mathbf{e}_{\mathrm{comp}}^{(\mathrm t)}$ with $\mathbf{e}_{\mathrm{comp}}^{(\mathrm s)}$ while sharing the level and position embeddings.

We concatenate the augmented embeddings along the feature dimension in a fixed component and level order:
\begin{equation}
\mathbf{Q}_{\mathrm{all}}=\texttt{Concat}\bigl(\overline{\mathbf{Q}}_1^{(\mathrm t)},\ldots,\overline{\mathbf{Q}}_L^{(\mathrm t)},\overline{\mathbf{Q}}_1^{(\mathrm s)},\ldots,\overline{\mathbf{Q}}_L^{(\mathrm s)}
\bigr)
\in\mathbb{R}^{N\times 2LD}.
\end{equation}
We then adopt a linear aggregation layer for these embeddings:
\begin{equation}
\mathbf{U}
=\texttt{LN}\!\left(
\texttt{Dropout}\!\left(
\texttt{GELU}\!\left(
\texttt{Dropout}(\mathbf{Q}_{\mathrm{all}})\mathbf{W}_u^\top
+\mathbf{1}_N\mathbf{b}_u^\top
\right)\right)\right),
\label{eq:token_aggregation}
\end{equation}
where $\mathbf{W}_u\in\mathbb{R}^{D\times 2LD}$ and $\mathbf{b}_u\in\mathbb{R}^{D\times 1}$ are trainable parameters.
The linear map uses the same weights for all rows, $\texttt{Dropout}$ applies element-wise dropout, and $\texttt{LN}$ normalizes the $D$ features within each row.

The gate fusion module $\texttt{GateFuse}$ projects the joint embedding and combines with the original embedding $\mathbf{R}$:
\begin{equation}
\begin{aligned}
\boldsymbol{\Delta}
&=\mathbf{U}\mathbf{W}_o^\top+\mathbf{1}_N\mathbf{b}_o^\top,\\
\mathbf{G}
&=\sigma\!\left(
[\mathbf{R}\Vert\boldsymbol{\Delta}]\mathbf{W}_g^\top
+\mathbf{1}_N\mathbf{b}_g^\top
\right),\\
\mathbf{R}'
&=\mathbf{R}+\mathbf{G}\odot\boldsymbol{\Delta},
\end{aligned}
\label{eq:gated_residual_fusion}
\end{equation}
where $\mathbf{W}_o\in\mathbb{R}^{D\times D}$, $\mathbf{W}_g\in\mathbb{R}^{D\times 2D}$, and $\mathbf{b}_o,\mathbf{b}_g\in\mathbb{R}^{D}$ are trainable parameters.
The concatenation $[\mathbf{R}\Vert\boldsymbol{\Delta}]\in\mathbb{R}^{N\times 2D}$ joins the two matrices along the feature dimension.
The sigmoid $\sigma$ produces $\mathbf{G}\in(0,1)^{N\times D}$, and $\odot$ denotes element-wise multiplication.

\noindent\textbf{Supervision Task.}
We can obtain the prediction results $\widehat{\mathbf{Y}} = F_\varphi(\mathbf{R}')$.
We respectively train the adaptation modules for each downstream dataset using paired input contexts and forecasting targets $(\mathbf{X},\mathbf{Y})$ in the training datasets.
Here, we do not introduce a new supervised loss function $\mathcal{L}_{\rm task}$ but adopt the TSFM-specific pre-trained loss. 
Taking the Chronos-Bolt as an example, we instantiate $\mathcal{L}_{\rm task}$ with its original native multi-quantile pinball loss:
\begin{equation}
\begin{aligned}
    \mathcal{L}_{\mathrm{task}}
    &=\frac{2}{N_{ q}}\sum_{k=1}^{N_{ q}}\sum_{h=1}^{H}
\rho_{\tau_k}\!\left(
{y}_h-\widehat{{y}}_h^{(k)}
\right),\\
\rho_{\tau_k}(e)&=\max\{\tau_k e,(\tau_k-1)e\},
\end{aligned}
\end{equation}
where $\{\tau_k | 0 < \tau_k < 1\}_{k=1}^{N_{ q}}$ denote the $N_{ q}$ native quantile levels.
${y}_h$ denotes the ground truth at the future time step $h$, and $\widehat{{y}}_h^{(k)}$ denotes the corresponding prediction at quantile level $\tau_k$. 
$\rho_{\tau_k}$ denotes the pinball loss.

\noindent \textbf{Training and Inference.}
In this stage, we freeze the encoder $F_{\theta}$ and decoder $F_{\varphi}$ modules in the TSFM and the learned codebook in Stage I.
We minimize $\mathcal{L}_{\mathrm{task}}$ with respect to the introduced parameters in hybrid aggregation and gate fusion for each downstream forecasting task.
At inference, we apply the Stage II forward process in Equation~\eqref{eq:stage2_overview} on each input context from the corresponding task to obtain the final prediction results.

%% file: table/main_backbone.tex
\begin{table*}[!htbp]
\centering
\caption{Overall forecasting results of \method across frozen TSFM backbones.
Results are averaged over five independent runs and forecasting horizons
$H \in \{96, 192, 336, 720\}$.
Electricity, Weather, and Traffic are included in the pre-training corpus
of TimesFM~2.5, and the corresponding results are omitted and marked with `--'.
We report relative error reductions \pctb{percentages}
compared with the corresponding zero-shot averages.
Full results are provided in Table~\ref{tab:backbone_appendix_main}.}
\label{tab:backbone_main}

\begin{small}
\setlength{\tabcolsep}{2.5pt}
\renewcommand{\arraystretch}{0.90}
\resizebox{\textwidth}{!}{%
\begin{tabular}{c|cc|cc|cc|cc|cc|cc|cc|cc}
\toprule
TSFM &
\multicolumn{4}{c|}{ChronosBolt-base} &
\multicolumn{4}{c|}{TimesFM~2.5} &
\multicolumn{4}{c|}{Moirai~2} &
\multicolumn{4}{c}{TTM~2} \\
\cmidrule{1-17}
Dataset &
\multicolumn{2}{c|}{Zero-shot} & \multicolumn{2}{c|}{\method} &
\multicolumn{2}{c|}{Zero-shot} & \multicolumn{2}{c|}{\method} &
\multicolumn{2}{c|}{Zero-shot} & \multicolumn{2}{c|}{\method} &
\multicolumn{2}{c|}{Zero-shot} & \multicolumn{2}{c}{\method} \\
\midrule
Metric &
MSE & MAE & MSE & MAE &
MSE & MAE & MSE & MAE &
MSE & MAE & MSE & MAE &
MSE & MAE & MSE & MAE \\
\midrule

ETTh1
& 0.4479 & 0.4185
& 0.4053\pctb{9.5\%} & 0.4056\pctb{3.1\%}
& 0.4793 & 0.4423
& 0.4480\pctb{6.5\%} & 0.4314\pctb{2.5\%}
& 0.4438 & 0.4147
& 0.3931\pctb{11.4\%} & 0.4028\pctb{2.9\%}
& 0.5492 & 0.5048
& 0.3822\pctb{30.4\%} & 0.4170\pctb{17.4\%} \\
\midrule

ETTh2
& 0.3666 & 0.3795
& 0.3116\pctb{15.0\%} & 0.3557\pctb{6.3\%}
& 0.4022 & 0.4094
& 0.3643\pctb{9.4\%} & 0.3977\pctb{2.9\%}
& 0.3884 & 0.3928
& 0.3728\pctb{4.0\%} & 0.3907\pctb{0.5\%}
& 0.4008 & 0.4291
& 0.2984\pctb{25.5\%} & 0.3600\pctb{16.1\%} \\
\midrule

ETTm1
& 0.4208 & 0.3829
& 0.3510\pctb{16.6\%} & 0.3734\pctb{2.5\%}
& 0.4290 & 0.4160
& 0.3820\pctb{11.0\%} & 0.3959\pctb{4.8\%}
& 0.5138 & 0.4425
& 0.3623\pctb{29.5\%} & 0.3775\pctb{14.7\%}
& 0.7426 & 0.5560
& 0.3593\pctb{51.6\%} & 0.3822\pctb{31.3\%} \\
\midrule

ETTm2
& 0.2905 & 0.3173
& 0.2658\pctb{8.5\%} & 0.3034\pctb{4.4\%}
& 0.3315 & 0.3412
& 0.2785\pctb{16.0\%} & 0.3240\pctb{5.0\%}
& 0.3569 & 0.3462
& 0.2638\pctb{26.1\%} & 0.3100\pctb{10.5\%}
& 0.3244 & 0.3659
& 0.2651\pctb{18.3\%} & 0.3200\pctb{12.5\%} \\
\midrule

Exchange
& 0.3712 & 0.4121
& 0.3508\pctb{5.5\%} & 0.3982\pctb{3.4\%}
& 0.4030 & 0.4269
& 0.3668\pctb{9.0\%} & 0.4102\pctb{3.9\%}
& 0.3795 & 0.4108
& 0.2517\pctb{33.7\%} & 0.3550\pctb{13.6\%}
& 0.4609 & 0.4677
& 0.4280\pctb{7.1\%} & 0.4352\pctb{6.9\%} \\
\midrule

Electricity
& 0.1588 & 0.2439
& 0.1528\pctb{3.8\%} & 0.2407\pctb{1.3\%}
& -- & -- & -- & --
& 0.1634 & 0.2475
& 0.1575\pctb{3.6\%} & 0.2434\pctb{1.7\%}
& 0.4892 & 0.5467
& 0.1638\pctb{66.5\%} & 0.2648\pctb{51.6\%} \\
\midrule

Weather
& 0.2631 & 0.2762
& 0.2282\pctb{13.3\%} & 0.2504\pctb{9.3\%}
& -- & -- & -- & --
& 0.2730 & 0.2753
& 0.2377\pctb{12.9\%} & 0.2639\pctb{4.1\%}
& 0.2215 & 0.2373
& 0.2047\pctb{7.6\%} & 0.2304\pctb{2.9\%} \\
\midrule

Traffic
& 0.4349 & 0.2627
& 0.3756\pctb{13.6\%} & 0.2326\pctb{11.5\%}
& -- & -- & -- & --
& 0.3908 & 0.2427
& 0.3834\pctb{1.9\%} & 0.2380\pctb{1.9\%}
& 0.3696 & 0.2437
& 0.3674\pctb{0.6\%} & 0.2410\pctb{1.1\%} \\

\bottomrule
\end{tabular}%
}
\end{small}
\end{table*}

%% file: section/04_experiment.tex
\section{Experiments}
\label{sec:experiments}

We conduct extensive experiments to answer the four research questions as follows.
\begin{itemize}[leftmargin=*,itemsep=0em]
    \item \textbf{RQ1:} Can \method enhance the TSFMs' forecasting and outperform existing parametric adaptation and knowledge augmentation approaches? \textit{(\textbf{Effectiveness})}
    \item \textbf{RQ2:} How does \method perform under backbone scaling and cross-model/cross-dataset transfer? \textit{(\textbf{Robustness})} 
    \item \textbf{RQ3:}  How does \method work? Which components play the key roles for the effectiveness of \method? \textit{(\textbf{Mechanism})}
    \item \textbf{RQ4:}  How does the efficiency of \method compare with that of the existing baseline methods? \textit{(\textbf{Efficiency})} 
\end{itemize}

\subsection{Experimental Setup}
\label{sec:exp-setup}

\textbf{Datasets.}
To guarantee fair comparison, we follow TS-RAG to construct the bank of input contexts $\mathcal{D}_{\rm pre}$.
We evaluate on eight widely-used multivariate forecasting benchmarks (ETTh1/2, ETTm1/2, Electricity, Exchange-rate, Traffic, Weather)~\cite{zhou2021informer}, covering diverse domains and sampling frequencies. 
More details in Table~\ref{tab:dataset}.

\noindent\textbf{Baselines.}
We apply our proposed \method to four representative TSFMs, including encoder--decoder Chronos-Bolt-base~\cite{ansari2024chronos}, decoder-only TimesFM~2.5~\cite{das2024decoder}, encoder-only Moirai~2~\cite{woo2024unified} and MLP-based TTM~2 ~\cite{ekambaram2024tiny}.
We compare \method with five representative baselines: RAFT~\cite{tire2024retrieval}, SARAF~\cite{zhou2026stationarity}, MEMTS~\cite{yu2026memts}, TS-RAG~\cite{ning2026ts}, and TS-Memory~\cite{lyu2026ts}. 
More details of these TSFMs and baselines are provided in Appendix~\ref{appendix_setup}.

\noindent\textbf{Implementations.}
The context length $T$ is set to~512 for both Stages.
To evaluate multi-scale prediction performance, we set the horizon length $H\in\{96,192,336,720\}$. 
The mean squared error (MSE) and mean absolute error (MAE) are adopted as the evaluation metrics. 
The residual level $L$, codebook size $K$, and alignment task coefficient $\lambda_{\rm align}$ are set to 2, 256, and 0.25 respectively.
The detailed hyperparameter investigations are provided in Appendix~\ref{sec_moredisscuison}.
In Stage I and II, training is iterated for 20 and 5 epochs respectively.
The two stages are optimized with AdamW with a fixed learning rate of \(10^{-4}\) and a weight decay of \(0.01\), and the batch size of $256$. 
All experiments are conducted on one NVIDIA A100-SXM4-40GB GPU. 

\input{table/main_baseline}

\input{table/main_transfer}

\subsection{RQ1: Effectiveness Study}
\label{sec:exp-main}
\noindent\textbf{Performance Across Backbones.}
Table~\ref{tab:backbone_main} reports the forecasting results of \method across four frozen TSFM backbones, averaged over five independent runs and four forecasting horizons.
\method reduces both MSE and MAE in all 29 evaluation settings under this horizon-averaged evaluation, consistently outperforming the zero-shot performance.
The average relative MSE reductions across the evaluated datasets are 10.7\%, 10.4\%, 15.4\%, and 26.0\% for ChronosBolt-base, TimesFM~2.5, Moirai~2, and TTM~2, respectively.
The magnitude of improvement varies across backbones and datasets.
For example, on Electricity, \method reduces MSE by 3.8\% and 3.6\% for ChronosBolt-base and Moirai~2, respectively, compared with 66.5\% for TTM~2.
These results demonstrate the effectiveness of \method across the evaluated frozen TSFMs, and  the improvements vary according to the backbones and downstream datasets.

\noindent\textbf{Comparison with Baselines.}
Table \ref{tab:main_baseline} compares \method with five representative baselines on ChronosBolt-base.
\method achieves the lowest average MSE or MAE results on seven of the eight datasets.
Compared with the best baseline in terms of MSE on each dataset, \method reduces MSE by 3.3\% on ETTh2, 6.7\% on Weather, and 9.0\% on Traffic.
These gains are consistent with the motivation for fine-grained knowledge augmentation.
Retrieving trend and seasonal patterns from separate residual codebooks enables temporal knowledge to be reused beyond its original combinations in individual contexts.
The task-adaptive adapter learns how to combine these patterns with the original TSFM representation under forecasting supervision.


\subsection{RQ2: Transferability Study} 

\begin{figure}[!t]
    \centering
    \includegraphics[width=\linewidth]{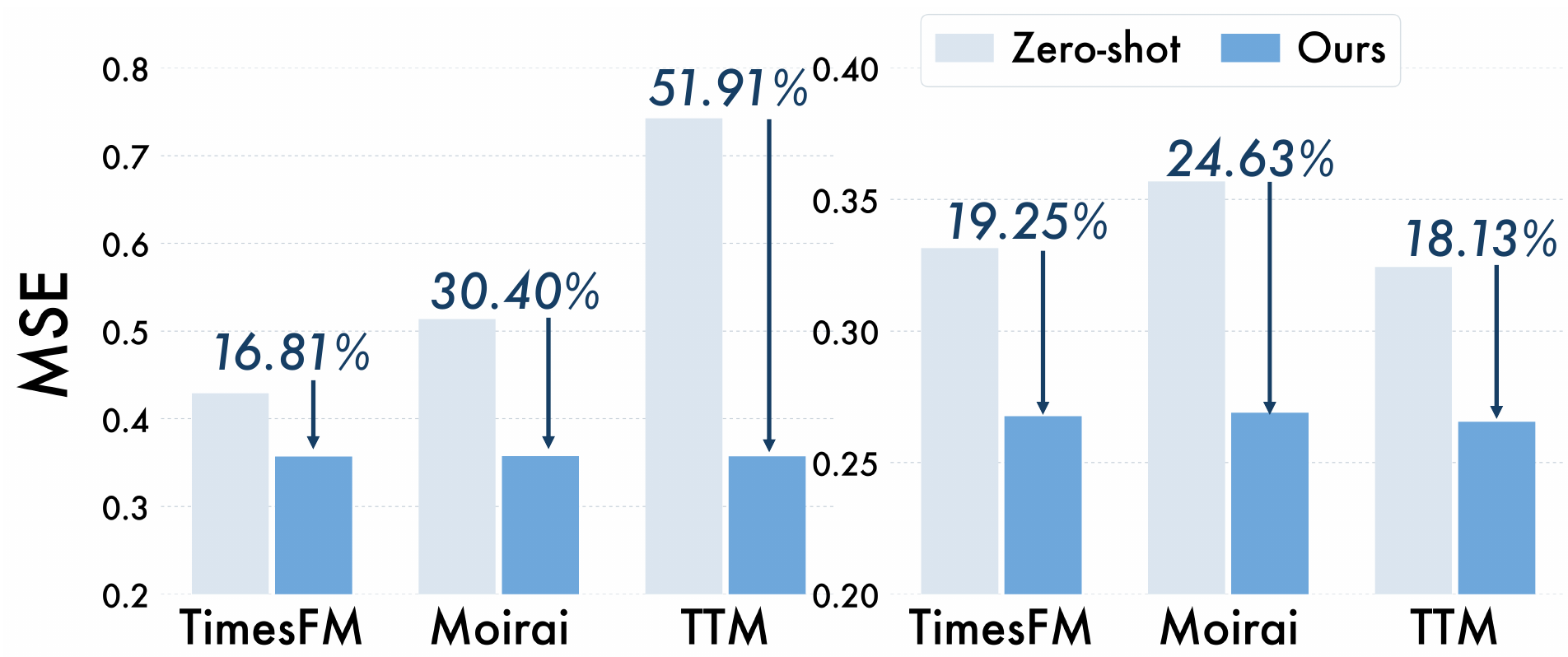}
    \caption{Cross-TSFM residual codebook transferability on ETTm1 (left) and ETTm2 (right).}
    \label{fig:transfer_2}
\end{figure}
\begin{figure}[!t]
    \centering
    \includegraphics[width=\linewidth]{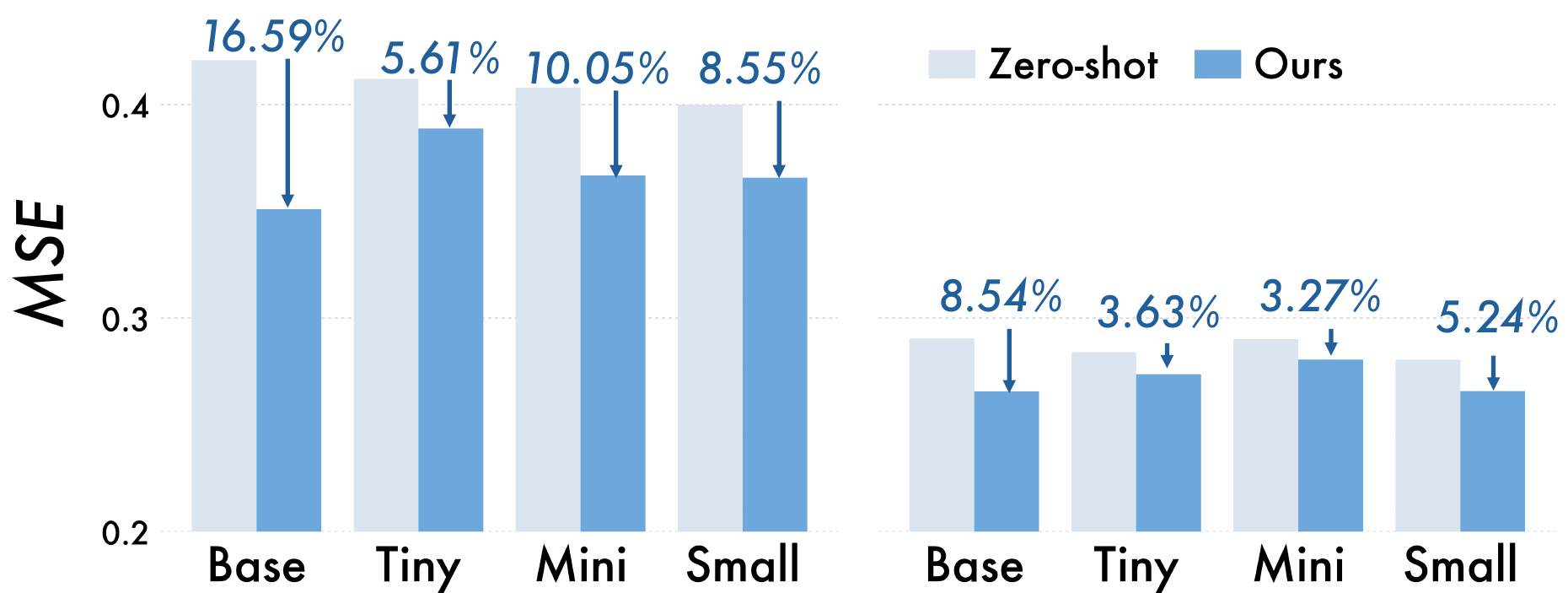}
    \caption{Residual codebook transferability across ChronosBolt model sizes on ETTm1 (left) and ETTm2 (right).}
    \label{fig:transfer_1}
\end{figure}

\noindent{\textbf{Cross-Dataset Adapter Transfer.}}
Table~\ref{tab:main_trans} evaluates the cross-dataset transferability of the Stage II adapter, including $\texttt{HyperAgg}$ and $\texttt{GateFuse}$, on ChronosBolt-base.
We train the adapter on the \textit{source} data and directly apply it to the \textit{target} dataset without further optimization.
We distinguish between \textit{within-domain transfer} between ETTm1 and ETTm2 and \textit{cross-domain transfer} from Weather to the ETTm datasets.
\method outperforms MEMTS in all four configurations, with relative MSE reductions of 2.69\%--4.77\% for within-domain transfer and 3.85\%--8.57\% for cross-domain transfer.
These results suggest that the adapter's learned use of shared trend and seasonal patterns can generalize beyond its training dataset.
We further evaluate pooled-source training using the four ETT datasets.
Because this pool includes the target datasets, these configurations assess multi-dataset adaptation.
Under this setting, \method reduces MSE by 13.45\% on ETTm1 and 8.87\% on ETTm2 compared with MEMTS, showing its effectiveness when the adapter is trained across multiple datasets.

\noindent{\textbf{Cross-TSFM Residual Codebook Transfer.}}
We evaluate whether the Stage I codebooks learned in one TSFM can be reused by other frozen TSFM backbones.
We transfer the learned trend and seasonal codebooks based on ChronosBolt-base to TimesFM~2.5, Moirai~2, and TTM~2.
For each target TSFM, we train a new Stage II adapter on for each downstream task, while keeping both the source codebooks and the target TSFM frozen.
We evaluate the forecasting performance on ETTm1 (left) and ETTm2 (right).
As shown in Figure~\ref{fig:transfer_2}, \method consistently outperforms the corresponding zero-shot results in all six evaluation settings.
The relative reductions range from 16.81\% to 51.91\% on ETTm1 and from 18.13\% to 24.63\% on ETTm2.
These results suggest that the temporal knowledge consolidated using the source TSFM remains useful for augmenting other forecasting backbones, supporting the reuse of Stage I pattern banks across different TSFMs.

\noindent{\textbf{Residual Codebook Transfer Across Model Sizes.}} 
We evaluate whether the learned residual codebooks in Stage I on ChronosBolt-base can support smaller backbones within the same model family.
The source codebooks remain fixed and are reused in Stage II with frozen ChronosBolt-Tiny (9M), Mini (21M), and Small (48M).
As shown in Figure~\ref{fig:transfer_1}, \method outperforms the corresponding zero-shot baselines at all four target sizes, reducing MSE by 5.61\%--16.59\% on ETTm1 and 3.27\%--8.54\% on ETTm2.
These gains suggest that recurring temporal patterns learned from the pretraining corpus remain useful across backbone scales, providing smaller models with input-relevant temporal regularities.



\begin{figure}[!t]
    \centering
    \includegraphics[width=\linewidth]{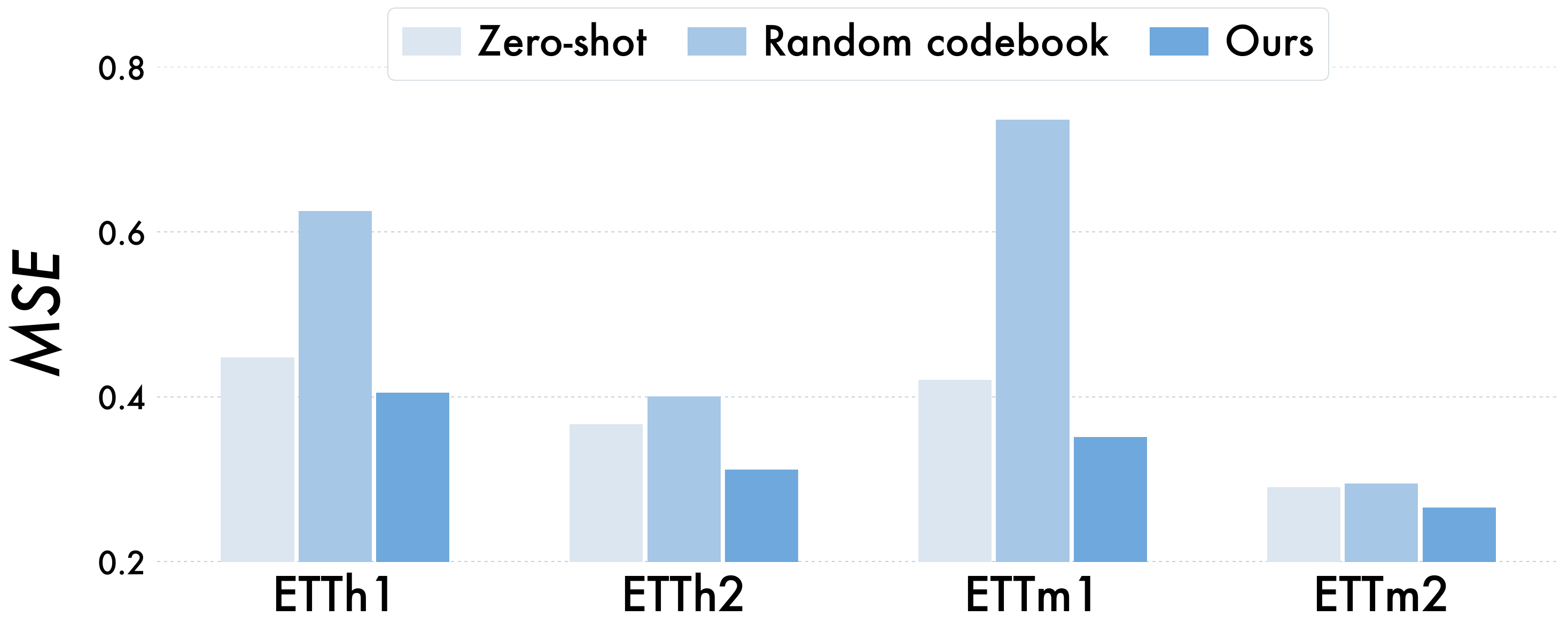}
    \caption{Forecasting with learned versus frozen random codebooks on four ETT datasets. Stage II adapters are trained separately for the two codebook settings.}
    \label{fig:knowledge}
\end{figure}
\begin{figure}[!t]
    \centering
    \includegraphics[width=\linewidth]{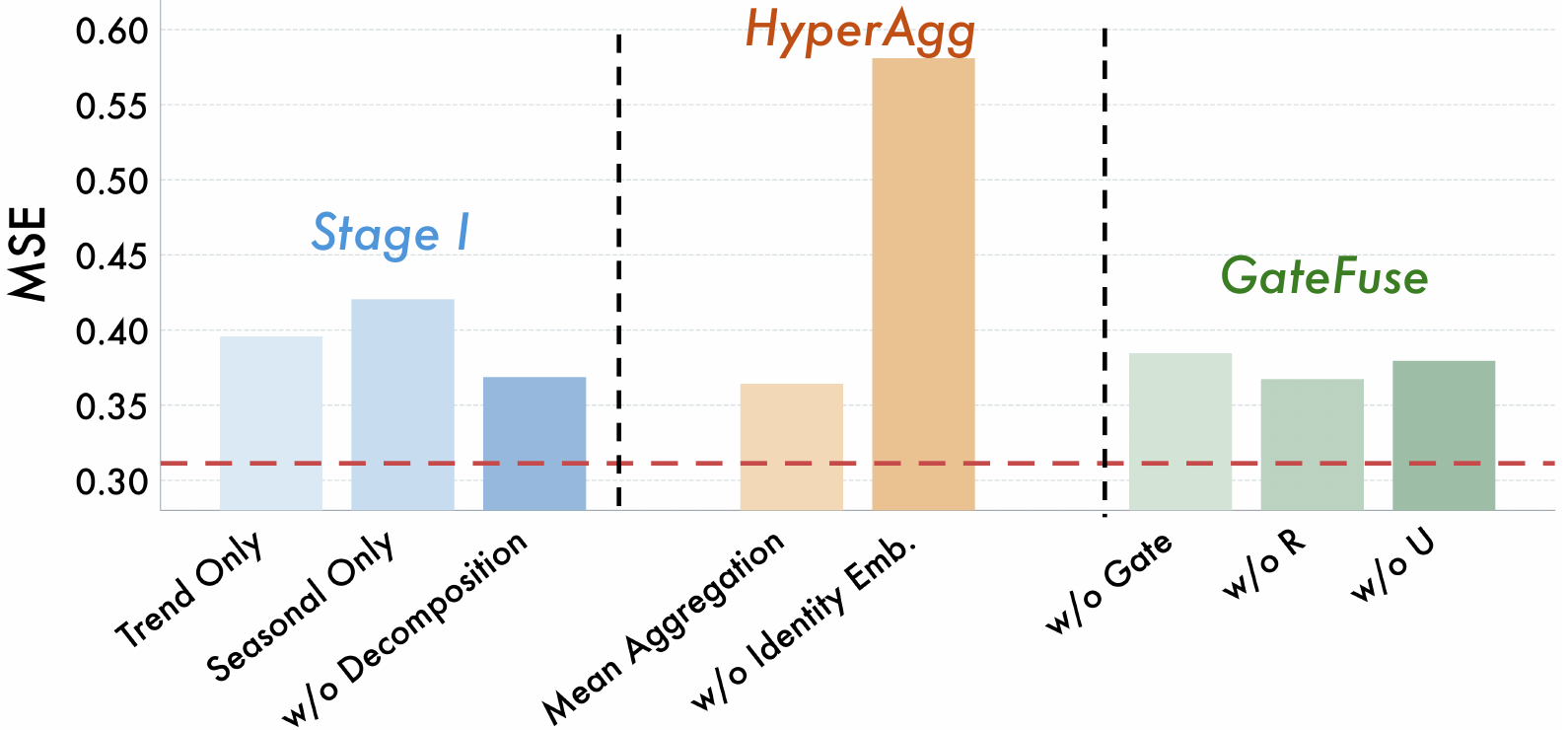}
    \caption{Ablation study on ETTh2, examining residual codebook construction, hybrid aggregation, and gate fusion. The dashed red line indicates the performance of \method.}
    \label{fig:ablation}
\end{figure}

\subsection{RQ3: Mechanism Study}

\noindent\textbf{Contribution of Residual Codebooks.}
We compare the learned Stage I codebooks with randomly initialized codebooks that remain frozen and train the Stage II adapter separately for each setting.
As shown in Figure~\ref{fig:knowledge}, random codebooks yield higher MSE than the learned codebooks and generally underperform the zero-shot backbone.
These results indicate that the gains of \method cannot be attributed solely to training the Stage II adapter, but the compact, reusable temporal patterns learned in Stage I also play an important role in improving forecasting accuracy.


\begin{figure}[!t]
    \centering
    \includegraphics[width=0.95\linewidth]{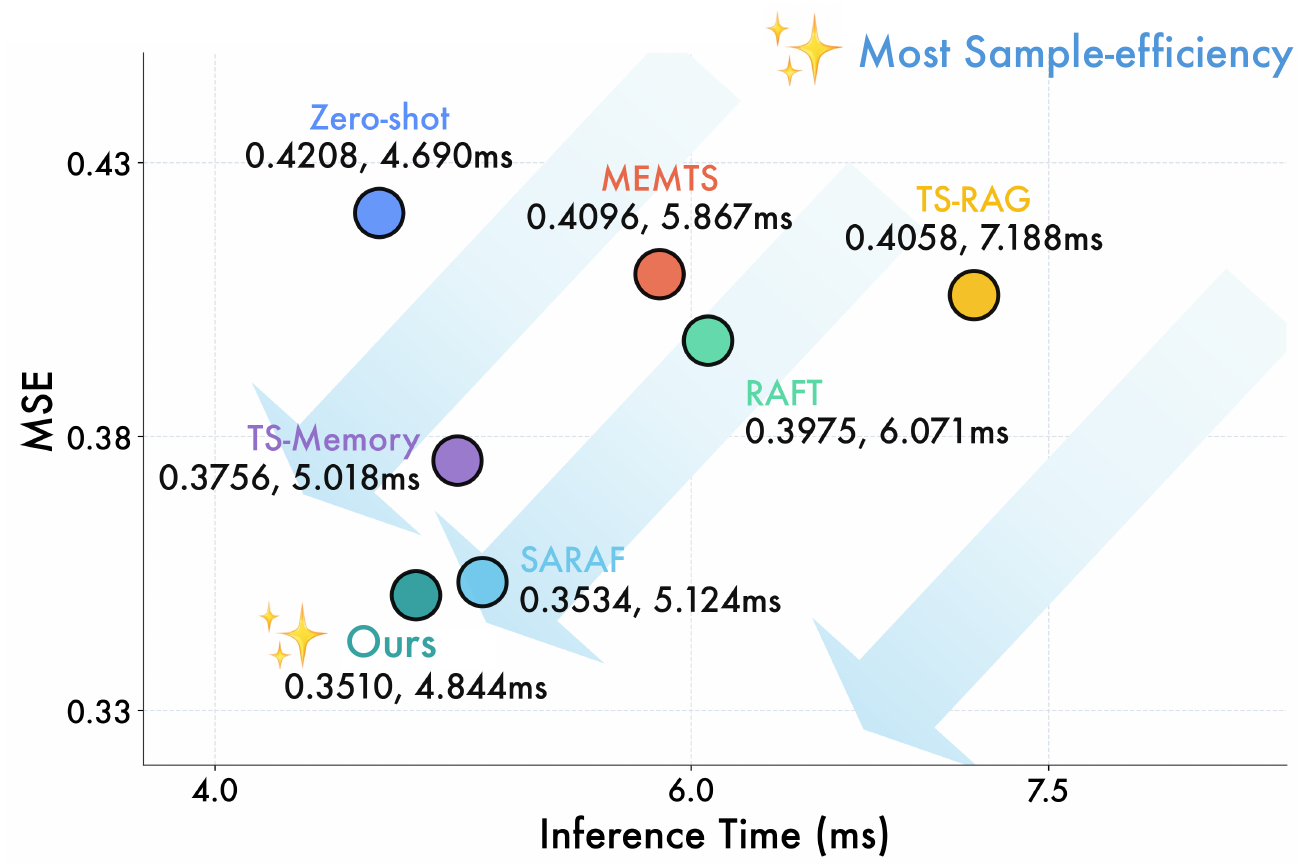}
    \caption{Forecasting accuracy and inference latency on ETTm1. Lower values are better on both axes.}
    \label{fig:efficiency}
\end{figure}

\noindent{\textbf{Ablation Study.}}
Figure~\ref{fig:ablation} compares \method with eight ablated variants on ETTh2. We can have the following observations.
\begin{itemize}[leftmargin=*]
    \item For $\texttt{Stage I}$, using only trend or seasonal knowledge reduces forecasting accuracy, suggesting that the two patterns provide complementary information.
    In \textit{w/o Decomposition}, we learn a single codebook from raw inputs, which also underperforms \method, supporting the effectiveness of the fine-grained knowledge augmentation.
    \item For \texttt{HyperAgg}, replacing learned parameters with mean operation increases MSE, indicating the benefit of learning how to combine the retrieved tokens.
    Removing identity embeddings causes the largest degradation among all variants.
    This highlights the importance of distinguishing tokens by component, residual level, and position when aggregating knowledge from multiple sources.
    \item For \texttt{GateFuse}, removing either the original TSFM embedding $\mathbf{R}$ or the aggregated knowledge  $\mathbf{U}$ degrades performance, supporting their complementary roles in forecasting. 
    The \textit{w/o Gate} variant replaces gated fusion with direct addition, i.e., $\mathbf{R}'=\mathbf{R}+\mathbf{U}$. Its higher MSE shows that learned gated fusion integrates the two representations more effectively than incorporating the two representations with equal weights.
\end{itemize}
Together, these results support both the organization of reusable temporal knowledge and its task-adaptive integration with the frozen TSFM.

\subsection{RQ4: Efficiency Study} 

We compare the forecasting accuracy and average inference latency of \method with the baselines on ETTm1 under the same evaluation setting, specified in Section~\ref{sec:exp-setup}.
Both MSE and inference latency are averaged over forecasting horizons $H\in\{96,192,336,720\}$.
As shown in Figure~\ref{fig:efficiency}, \method achieves the lowest MSE of 0.3510 while maintaining an inference latency of 4.844\,ms.
Compared with zero-shot forecasting, it reduces MSE by 16.6\% with only a 3.3\% increase in latency, indicating that the accuracy gain comes with limited additional inference cost.
\method is approximately $1.25\times$ faster than RAFT and $1.48\times$ faster than TS-RAG, while reducing MSE by 11.7\% and 13.5\%, respectively. It also outperforms TS-Memory and MEMTS in both MSE and inference latency.
This favorable trade-off is consistent with our design: Stage I consolidates temporal knowledge into compact pattern banks, so downstream retrieval does not require accessing the original pre-training windows.
Stage II then aggregates the selected pattern tokens and incorporates them through a lightweight gated adapter.


\begin{figure}[!t]
    \centering
    \includegraphics[width=\linewidth]{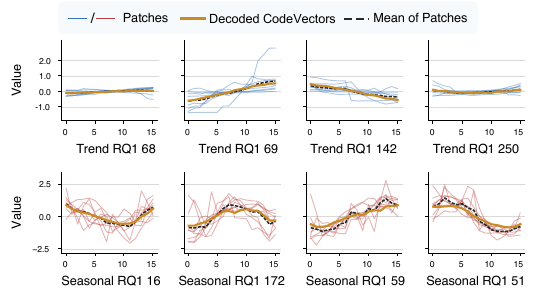}
    \caption{
Representative first-level trend and seasonal code vectors learned using ChronosBolt-base.
Thin blue/red curves show assigned patches from distinct sources, orange curves show decoded code vectors, and black dashed curves show the patch means.
}
    \label{fig:vis1}
\end{figure}
\begin{figure}[!t]
    \centering
    \includegraphics[width=\linewidth]{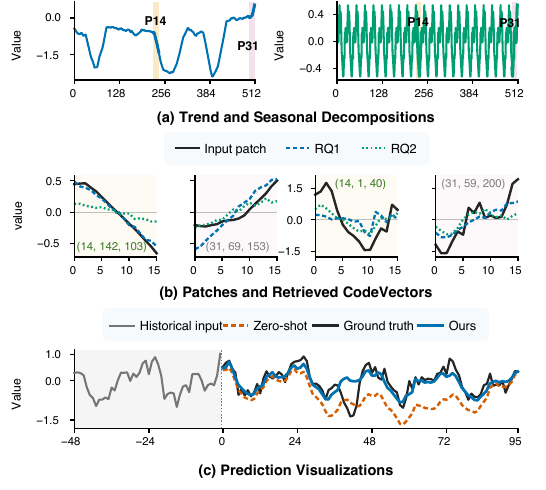}
    \caption{ Stage II case study on ETTh2 using ChronosBolt-base.
    (a) Trend and seasonal inputs.
    (b) Highlighted patches and decoded codebook entries in each residual levels. Tuples denote (patch index, RQ1 ID, RQ2 ID).
    (c) Prediction comparison.
}
    \label{fig:vis2}
\end{figure}




\subsection{Case Study}

\noindent\textbf{Codebook Interpretation.}
Figure~\ref{fig:vis1} illustrates the local temporal structures associated with representative first-level code vectors.
The trend vectors exhibit different shapes, with RQ1 \texttt{68} and \texttt{250} corresponding to relatively small local changes, and \texttt{69} and \texttt{142} to upward and downward movements, respectively.
The seasonal vectors represent different local waveforms and phase positions, including trough-like and crest-like structures.
Although individual patches vary in amplitude and local fluctuations, the decoded code vectors generally follow their average shapes.
These examples suggest that the codebooks consolidate common local structures across historical contexts while accommodating variation among the assigned patches.


\noindent\textbf{Pattern Retrieval and Forecasting Behavior.}
Figure~\ref{fig:vis2} illustrates code selection and forecasting behavior for a selected ETTh2 sample.
The trend patches \texttt{P14} and \texttt{P31} exhibit downward and upward movements, respectively, which are reflected in their selected RQ1 code vectors.
The seasonal vectors approximate broad local variations but smooth several extrema.
These examples illustrate the temporal structures associated with the selected code vectors.
Stage II retrieves these reusable patterns from the codebooks learned in Stage I and integrates them with the frozen TSFM representations to augment forecasting.
In Figure~\ref{fig:vis2}(c), the zero-shot forecast underestimates the observed level over much of the later horizon, while \method follows it more closely.
For this window, MSE decreases from 0.5004 to 0.1126 and MAE from 0.5863 to 0.2279.
This case illustrates forecasting behavior but may not isolate the contribution of individual codes.

%% file: table/main_baseline.tex
\begin{table*}[!htbp]
\centering
\caption{Forecasting results comparison on ChronosBolt-base.
Results are averaged over forecasting horizons $H \in \{96,192,336,720\}$.
The best results are highlighted in \firstres{purple},
and the second-best underlined in \secondres{green};
ties are determined at the displayed precision.
We report relative error reductions \pctb{percentages}
and increases \pctr{percentages} compared with zero-shot forecasting,
computed from the displayed average results.
OOM: out of memory.
Full results are provided in Table~\ref{tab:chronosbolt_main_appendix}.}
\label{tab:main_baseline}

\setlength{\tabcolsep}{2.5pt}
\renewcommand{\arraystretch}{0.90}
\resizebox{\textwidth}{!}{%
\begin{tabular}{c|cc|cc|cc|cc|cc|cc|cc}
\toprule
Dataset &
\multicolumn{2}{c|}{Zero-shot} &
\multicolumn{2}{c|}{RAFT} &
\multicolumn{2}{c|}{TS-RAG} &
\multicolumn{2}{c|}{TS-Memory} &
\multicolumn{2}{c|}{SARAF} &
\multicolumn{2}{c|}{MEMTS} &
\multicolumn{2}{c}{\method} \\
\midrule
Metric &
MSE & MAE &
MSE & MAE &
MSE & MAE &
MSE & MAE &
MSE & MAE &
MSE & MAE &
MSE & MAE \\
\midrule

ETTh1
& 0.4479 & 0.4185
& 0.4368\pctb{2.5\%} & 0.4176\pctb{0.2\%}
& 0.4423\pctb{1.3\%} & 0.4214\pctr{0.7\%}
& 0.4310\pctb{3.8\%} & 0.4199\pctr{0.3\%}
& \secondres{0.4095}\pctb{8.6\%} & 0.4276\pctr{2.2\%}
& 0.4332\pctb{3.3\%} & \secondres{0.4093}\pctb{2.2\%}
& \firstres{0.4053}\pctb{9.5\%} & \firstres{0.4056}\pctb{3.1\%} \\

\midrule
ETTh2
& 0.3666 & 0.3795
& 0.3574\pctb{2.5\%} & 0.3776\pctb{0.5\%}
& 0.3616\pctb{1.4\%} & 0.3853\pctr{1.5\%}
& 0.3634\pctb{0.9\%} & 0.3790\pctb{0.1\%}
& 0.3411\pctb{7.0\%} & 0.3918\pctr{3.2\%}
& \secondres{0.3224}\pctb{12.1\%} & \secondres{0.3559}\pctb{6.2\%}
& \firstres{0.3116}\pctb{15.0\%} & \firstres{0.3557}\pctb{6.3\%} \\

\midrule
ETTm1
& 0.4208 & 0.3829
& 0.3975\pctb{5.5\%} & 0.3768\pctb{1.6\%}
& 0.4058\pctb{3.6\%} & 0.3813\pctb{0.4\%}
& 0.3756\pctb{10.7\%} & \firstres{0.3722}\pctb{2.8\%}
& \secondres{0.3534}\pctb{16.0\%} & 0.3816\pctb{0.3\%}
& 0.4096\pctb{2.7\%} & 0.3849\pctr{0.5\%}
& \firstres{0.3510}\pctb{16.6\%} & \secondres{0.3734}\pctb{2.5\%} \\

\midrule
ETTm2
& 0.2905 & 0.3173
& 0.2810\pctb{3.3\%} & \secondres{0.3145}\pctb{0.9\%}
& 0.2829\pctb{2.6\%} & 0.3181\pctr{0.3\%}
& 0.2751\pctb{5.3\%} & 0.3163\pctb{0.3\%}
& \firstres{0.2520}\pctb{13.3\%} & 0.3153\pctb{0.6\%}
& 0.2950\pctr{1.5\%} & 0.3247\pctr{2.3\%}
& \secondres{0.2658}\pctb{8.5\%} & \firstres{0.3034}\pctb{4.4\%} \\

\midrule
Exchange
& 0.3712 & 0.4121
& 0.3687\pctb{0.7\%} & 0.4085\pctb{0.9\%}
& \secondres{0.3669}\pctb{1.2\%} & \secondres{0.4079}\pctb{1.0\%}
& 0.3980\pctr{7.2\%} & 0.4314\pctr{4.7\%}
& 0.3945\pctr{6.3\%} & 0.4192\pctr{1.7\%}
& 0.3775\pctr{1.7\%} & 0.4116\pctb{0.1\%}
& \firstres{0.3508}\pctb{5.5\%} & \firstres{0.3982}\pctb{3.4\%} \\

\midrule
Weather
& 0.2631 & 0.2762
& 0.2539\pctb{3.5\%} & \secondres{0.2680}\pctb{3.0\%}
& 0.2615\pctb{0.6\%} & 0.2794\pctr{1.2\%}
& \secondres{0.2446}\pctb{7.0\%} & 0.2824\pctr{2.2\%}
& 0.2454\pctb{6.7\%} & 0.2939\pctr{6.4\%}
& 0.2575\pctb{2.1\%} & 0.2914\pctr{5.5\%}
& \firstres{0.2282}\pctb{13.3\%} & \firstres{0.2504}\pctb{9.3\%} \\

\midrule
Electricity
& 0.1588 & 0.2439
& 0.1564\pctb{1.5\%} & 0.2443\pctr{0.2\%}
& 0.1567\pctb{1.3\%} & 0.2445\pctr{0.2\%}
& 0.1573\pctb{0.9\%} & \secondres{0.2434}\pctb{0.2\%}
& \secondres{0.1529}\pctb{3.7\%} & 0.2538\pctr{4.1\%}
& OOM & OOM
& \firstres{0.1528}\pctb{3.8\%} & \firstres{0.2407}\pctb{1.3\%} \\

\midrule
Traffic
& 0.4349 & 0.2627
& 0.4265\pctb{1.9\%} & \secondres{0.2608}\pctb{0.7\%}
& 0.4267\pctb{1.9\%} & 0.2609\pctb{0.7\%}
& 0.4273\pctb{1.7\%} & \secondres{0.2608}\pctb{0.7\%}
& \secondres{0.4127}\pctb{5.1\%} & 0.2943\pctr{12.0\%}
& OOM & OOM
& \firstres{0.3756}\pctb{13.6\%} & \firstres{0.2326}\pctb{11.5\%} \\

\bottomrule
\end{tabular}%
}
\end{table*}

%% file: table/main_transfer.tex
\begin{table}[!t]
\centering

\caption{Cross-dataset adapter transferability on ChronosBolt-base.
We report the relative performance increase \pctb{percentages} compared to MEMTS. 
ETT$_{\text{pool}}$: pooled four ETT datasets.
}
\label{tab:main_trans}

{\small
\setlength{\tabcolsep}{2pt}
\renewcommand{\arraystretch}{1}

\scalebox{1.00}{%
\begin{tabular}{c|cc|cc}
\toprule
\multicolumn{1}{c|}{} & \multicolumn{2}{c|}{{MEMTS}} & \multicolumn{2}{c}{{\method}} \\
\midrule
\multicolumn{1}{c|}{Source$\rightarrow$Target} &
\multicolumn{1}{c}{MSE} &
\multicolumn{1}{c|}{MAE} &
\multicolumn{1}{c}{MSE} &
\multicolumn{1}{c}{MAE} \\
\midrule

Weather$\rightarrow$ ETTm1
&0.4203 & 0.3974  & 0.3843\pctb{8.57\%} & 0.3948\pctb{0.65\%} \\
\midrule

Weather$\rightarrow$ ETTm2
& 0.3113 & 0.3620 & 0.2993\pctb{3.85\%} & 0.3247\pctb{10.30\%} \\
\midrule


ETTm2 $\rightarrow$ ETTm1 
& 0.4190 & 0.3834  & 0.3990\pctb{4.77\%} & 0.3782\pctb{1.36\%}\\
\midrule

ETTm1 $\rightarrow$ ETTm2
& 0.2941 & 0.3367 & 0.2862\pctb{2.69\%} & 0.3187\pctb{5.35\%} \\
\midrule


ETT$_{\text{pool}}$ $\rightarrow$ ETTm1
& 0.4179 & 0.3877  & 0.3617\pctb{13.45\%} & 0.3741\pctb{3.51\%} \\
\midrule

ETT$_{\text{pool}}$ $\rightarrow$ ETTm2
& 0.2919 & 0.3273  & 0.2660\pctb{8.87\%} & 0.3059\pctb{6.54\%} \\

\bottomrule 
\end{tabular}}%
}
\end{table} 

%% file: section/06_conclusion.tex
\section{Conclusion and Future Work}

In this paper, we propose \method, a fine-grained knowledge augmentation framework for frozen TSFMs.
\method consolidates trend and seasonal knowledge from pre-training corpora into shared residual codebooks and learns to aggregate and fuse retrieved patterns for downstream forecasting.
This design enables temporal knowledge to be reused beyond its original combinations in individual contexts, while separating shared knowledge consolidation from task-adaptive knowledge use.
Experiments on eight real-world datasets with four TSFM backbones demonstrate consistent improvements over the corresponding zero-shot baselines.

\noindent\textbf{Limitations and Future Work.}
Nevertheless, the fixed trend--seasonal decomposition imposes a predefined organization of temporal knowledge, and the frozen pattern banks do not incorporate new knowledge during downstream adaptation.
Future work will explore data-driven pattern organization and continual updates to the shared banks to support knowledge reuse in evolving domains.

%% file: appendix/0_appendix_list.tex
\input{appendix/c_moresetting.tex}

\input{appendix/e_moredisscuision}

\input{table/app/backbone_append}

\input{table/app/chrons_bolt_baseline}

%% file: appendix/c_moresetting.tex
\section{Experimental Details}
\label{appendix_setup}

\subsection{Details of TSFMs and baselines}\label{appendix_baseline}
In this appendix, we provide detailed descriptions of TSFMs and baselines.

\noindent\textbf{Time Series Foundation Models.}
\begin{itemize}[leftmargin=*,itemsep=1em,parsep=1.5pt]
   \item \textbf{{Chronos-Bolt}~\citep{ansari2024chronos}}:
       Chronos-Bolt is a patch-based variant of Chronos.
       It encodes non-overlapping patches with a T5 encoder--decoder and emits a multi-quantile forecast in one forward pass, avoiding the slow token-by-token decoding of the original Chronos models.
       {\url{https://huggingface.co/autogluon/chronos-bolt-base}}
       \item \textbf{{TimesFM~2.5}~\citep{das2024decoder}}:
       TimesFM~2.5 is a decoder-only foundation model that forecasts future patches autoregressively.
       Compared with earlier TimesFM releases, it uses 200M parameters, supports a much longer context, adds a continuous quantile head, and no longer requires a frequency indicator at inference.
       {\url{https://huggingface.co/google/timesfm-2.5-200m-pytorch}}.
       \item \textbf{{Moirai}~\citep{woo2024unified}}:
       Moirai is a masked-encoder Transformer trained as a universal probabilistic forecaster.
       Any-variate attention flattens an arbitrary number of channels into one sequence, and multi-patch-size projections handle different sampling frequencies.
       {\url{https://huggingface.co/Salesforce/moirai-2.0-R-small}}.
       \item \textbf{{TTM}~\citep{ekambaram2024tiny}}:
       TTM is a compact TSMixer-based foundation model.
       Adaptive patching and resolution prefix tuning allow it to pre-train across heterogeneous frequencies with about 1M parameters.
       {\url{https://huggingface.co/ibm-research/ttm-research-r2}}.
\end{itemize}

\noindent\textbf{Baselines.}
\begin{itemize}[leftmargin=*,itemsep=1em,parsep=1.5pt]
    \item \textbf{{RAFT}~\citep{tire2024retrieval}}:
       RAFT is a retrieval-augmented forecaster that looks up historical patches whose past most closely matches the query.
       Similarity is measured by Pearson correlation after multi-period downsampling, and the futures of the retrieved patches are aggregated through a temperature-scaled softmax before a shallow MLP produces the final forecast.
       \url{https://github.com/archon159/RAFT}
       \item \textbf{{SARAF}~\citep{zhou2026stationarity}}:
       SARAF is a stationarity-aware retrieval framework built on the observation that similar pasts need not imply similar futures.
       It first forms a time-aligned candidate pool, then applies MMR to trade off relevance against diversity according to a dataset-level stationarity score, and finally aggregates the retrieved futures with a stationarity-conditioned Gaussian kernel.
       \url{https://github.com/ShiqiaoZhou/SARAF}
       \item \textbf{{TS-RAG}~\citep{ning2026ts}}:
       TS-RAG is a retrieval-augmented generation framework for time series foundation models.
       A retriever indexes historical contexts by foundation-model embeddings and returns the top-$k$ neighbors together with their realized futures, after which an Adaptive Retrieval Mixer assigns importance scores and fuses the retrieved patterns with the backbone representation.
       \url{https://github.com/UConn-DSIS/TS-RAG}
       \item \textbf{{TS-Memory}~\citep{lyu2026ts}}:
       TS-Memory is a plug-and-play memory adapter for time series foundation models.
       An offline $k$NN teacher first synthesizes confidence-aware quantile targets from retrieved futures, and a lightweight memory module distills this distributional correction so that it can be interpolated with the backbone forecast at test time.
       \url{https://github.com/sisuolv/TS-Memory}
       \item \textbf{{MEMTS}~\citep{yu2026memts}}:
       MEMTS adapts a foundation model by internalizing domain knowledge into parameterized memory.
       A Knowledge Persistence Module compresses recurring trends and seasonal patterns into a small set of learnable prototypes, and an Adaptive Fusion module gates these memory tokens against the backbone forecast.
       \url{https://github.com/ulrica66666/TS-MEMTS}
\end{itemize}

%% file: appendix/e_moredisscuision.tex
\section{Hyperparameter Investigation}
\label{sec_moredisscuison}
\input{table/app/dataset} 

\begin{figure}[!t]
    \centering
    \includegraphics[width=\linewidth]{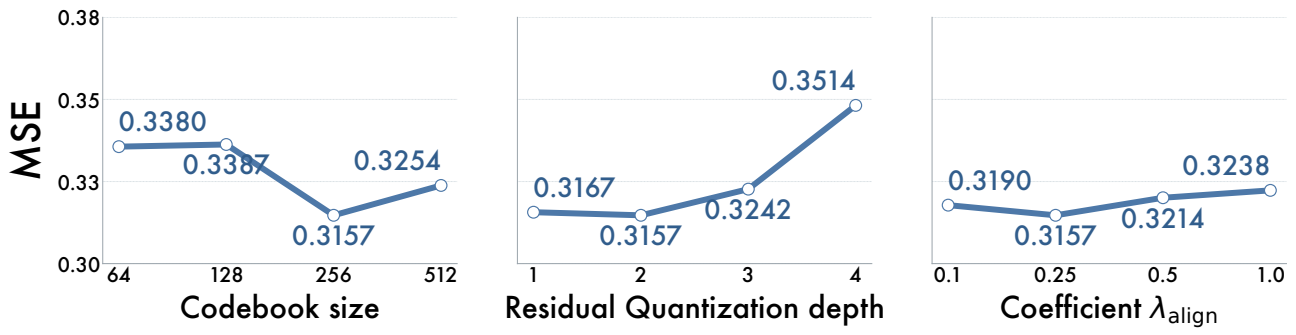}
    \caption{Sensitivity study of \method on ETTh2.}
    \label{fig:sensitvely}
\end{figure}

Figure~\ref{fig:sensitvely} analyzes the number of code vectors per level (64/ 128/ 256/ 512), different residual-quantization levels (1/ 2/ 3/ 4) and different coefficient values $\lambda_{\mathrm{align}}$ for \method on Etth2 (forecasting horizon $H = 336$).
Increasing $K$ from 64 to 256 improves forecasting accuracy, while further increasing it to 512 raises MSE, suggesting that a larger codebook size does not necessarily provide more useful predictive information.
For residual depth, $L=2$ outperforms $L=1$, whereas deeper residual levels degrade performance,
suggesting limited benefits from incorporating finer residual information in this setting.
The lowest MSE is achieved at $\lambda_{\mathrm{align}}=0.25$, supporting a balance between codebook alignment and reconstruction.
These results support our default configuration of $K=256$, $L=2$, and $\lambda_{\mathrm{align}}=0.25$.


%% file: table/app/dataset.tex

\begin{table}[!t]
  \caption{Detailed statistics of all benchmarking datasets.}
  \label{tab:dataset}
  \centering
  \footnotesize
  \setlength{\tabcolsep}{3pt}
  \renewcommand{\arraystretch}{1.1}
  \resizebox{\columnwidth}{!}{%
    \begin{tabular}{@{}lcccccc@{}}
      \toprule
      Dataset & Dim. & Freq. & Periodicity
      & \# Train/Val/Test & Time Span & Domain \\
      \midrule
      ETTm1 & 7 & 15 min & 96
      & (34465, 11521, 11521) & 2016.07--2018.06 & Temperature \\
      ETTm2 & 7 & 15 min & 96
      & (34465, 11521, 11521) & 2016.07--2018.06 & Temperature \\
      ETTh1 & 7 & 1 hour & 24
      & (8545, 2881, 2881) & 2016.07--2018.06 & Temperature \\
      ETTh2 & 7 & 1 hour & 24
      & (8545, 2881, 2881) & 2016.07--2018.06 & Temperature \\
      Electricity & 321 & 1 hour & 24
      & (18317, 2633, 5261) & 2016.07--2019.07 & Electricity \\
      Exchange-rate & 8 & 1 day & 7
      & (5216, 761, 1518) & 1990.01--2010.10 & Finance \\
      Traffic & 862 & 1 hour & 24
      & (12185, 1757, 3509) & 2016.07--2018.07 & Transportation \\
      Weather & 21 & 10 min & 144
      & (36792, 5271, 10540) & 2020.01--2021.01 & Weather \\
      \bottomrule
    \end{tabular}%
  }
\end{table}

%% file: table/app/backbone_append.tex

\begingroup
\newcommand{\tfkaresult}[2]{%
  #1\textcolor{gray}{\scriptsize$\pm$#2}%
}

\begin{table*}[!htpb]
\centering
\caption{Overall forecasting results of \method across frozen TSFM backbones.
Results at each forecasting horizon are averaged over five independent runs.
The Avg rows summarize performance across forecasting horizons
$H \in \{96, 192, 336, 720\}$.
Electricity, Weather, and Traffic are included in the pre-training corpus
of TimesFM~2.5, and the corresponding results are omitted and marked with `--'.
The $\Delta$ rows report relative error reductions
\pctb{percentages} compared with the corresponding zero-shot averages.}
\label{tab:backbone_appendix_main}

\begin{small}
\setlength{\tabcolsep}{2pt}
\resizebox{\textwidth}{!}{%
\begin{tabular}{c|c|cc|cc|cc|cc|cc|cc|cc|cc}
\toprule
\multicolumn{2}{c|}{TSFM} &
\multicolumn{4}{c|}{ChronosBolt-base} &
\multicolumn{4}{c|}{TimesFM~2.5} &
\multicolumn{4}{c|}{Moirai~2} &
\multicolumn{4}{c}{TTM~2} \\
\midrule
\multicolumn{2}{c|}{Dataset} &
\multicolumn{2}{c|}{Zero-shot} & \multicolumn{2}{c|}{\method} &
\multicolumn{2}{c|}{Zero-shot} & \multicolumn{2}{c|}{\method} &
\multicolumn{2}{c|}{Zero-shot} & \multicolumn{2}{c|}{\method} &
\multicolumn{2}{c|}{Zero-shot} & \multicolumn{2}{c}{\method} \\
\midrule
\multicolumn{2}{c|}{Metric} &
MSE & MAE & MSE & MAE &
MSE & MAE & MSE & MAE &
MSE & MAE & MSE & MAE &
MSE & MAE & MSE & MAE \\
\midrule

\multirow{6}{*}{\rotatebox{90}{ETTh1}}
& 96
& 0.3849 & 0.3798
& \tfkaresult{0.3570}{0.0087} & \tfkaresult{0.3720}{0.0047}
& 0.4211 & 0.4009
& \tfkaresult{0.3920}{0.0072} & \tfkaresult{0.3901}{0.0024}
& 0.3838 & 0.3803
& \tfkaresult{0.3572}{0.0075} & \tfkaresult{0.3733}{0.0044}
& 0.5039 & 0.4723
& \tfkaresult{0.3376}{0.0106} & \tfkaresult{0.3812}{0.0057} \\

& 192
& 0.4417 & 0.4122
& \tfkaresult{0.3978}{0.0112} & \tfkaresult{0.3964}{0.0064}
& 0.4721 & 0.4323
& \tfkaresult{0.4406}{0.0164} & \tfkaresult{0.4209}{0.0039}
& 0.4257 & 0.4071
& \tfkaresult{0.3904}{0.0273} & \tfkaresult{0.3969}{0.0151}
& 0.5415 & 0.4948
& \tfkaresult{0.3623}{0.0136} & \tfkaresult{0.3994}{0.0062} \\

& 336
& 0.4814 & 0.4331
& \tfkaresult{0.4208}{0.0128} & \tfkaresult{0.4131}{0.0058}
& 0.5098 & 0.4548
& \tfkaresult{0.4578}{0.0233} & \tfkaresult{0.4394}{0.0112}
& 0.4600 & 0.4237
& \tfkaresult{0.4068}{0.0343} & \tfkaresult{0.4092}{0.0216}
& 0.5599 & 0.5104
& \tfkaresult{0.3833}{0.0119} & \tfkaresult{0.4181}{0.0058} \\

& 720
& 0.4836 & 0.4489
& \tfkaresult{0.4456}{0.0102} & \tfkaresult{0.4408}{0.0037}
& 0.5140 & 0.4813
& \tfkaresult{0.5016}{0.0407} & \tfkaresult{0.4752}{0.0255}
& 0.5055 & 0.4475
& \tfkaresult{0.4181}{0.1362} & \tfkaresult{0.4320}{0.0770}
& 0.5916 & 0.5418
& \tfkaresult{0.4456}{0.0059} & \tfkaresult{0.4694}{0.0033} \\

\cmidrule{2-18}
& Avg
& 0.4479 & 0.4185
& \tfkaresult{0.4053}{0.0105} & \tfkaresult{0.4056}{0.0049}
& 0.4793 & 0.4423
& \tfkaresult{0.4480}{0.0219} & \tfkaresult{0.4314}{0.0097}
& 0.4438 & 0.4147
& \tfkaresult{0.3931}{0.0304} & \tfkaresult{0.4028}{0.0169}
& 0.5492 & 0.5048
& \tfkaresult{0.3822}{0.0102} & \tfkaresult{0.4170}{0.0047} \\

& $\Delta$
& -- & -- & \pctb{9.51\%} & \pctb{3.08\%}
& -- & -- & \pctb{6.53\%} & \pctb{2.46\%}
& -- & -- & \pctb{11.42\%} & \pctb{2.87\%}
& -- & -- & \pctb{30.41\%} & \pctb{17.39\%} \\
\midrule

\multirow{6}{*}{\rotatebox{90}{ETTh2}}
& 96
& 0.2895 & 0.3251
& \tfkaresult{0.2424}{0.0197} & \tfkaresult{0.2982}{0.0124}
& 0.3263 & 0.3515
& \tfkaresult{0.2966}{0.0045} & \tfkaresult{0.3413}{0.0015}
& 0.3059 & 0.3337
& \tfkaresult{0.2943}{0.0049} & \tfkaresult{0.3323}{0.0037}
& 0.3250 & 0.3814
& \tfkaresult{0.2309}{0.0190} & \tfkaresult{0.3064}{0.0113} \\

& 192
& 0.3598 & 0.3695
& \tfkaresult{0.2812}{0.0309} & \tfkaresult{0.3310}{0.0170}
& 0.3994 & 0.3979
& \tfkaresult{0.3595}{0.0174} & \tfkaresult{0.3873}{0.0123}
& 0.3847 & 0.3834
& \tfkaresult{0.3642}{0.0053} & \tfkaresult{0.3784}{0.0043}
& 0.3970 & 0.4267
& \tfkaresult{0.2716}{0.0382} & \tfkaresult{0.3395}{0.0198} \\

& 336
& 0.4029 & 0.4025
& \tfkaresult{0.3157}{0.0306} & \tfkaresult{0.3633}{0.0155}
& 0.4335 & 0.4311
& \tfkaresult{0.3831}{0.0268} & \tfkaresult{0.4154}{0.0235}
& 0.4159 & 0.4138
& \tfkaresult{0.3786}{0.0176} & \tfkaresult{0.3993}{0.0111}
& 0.4164 & 0.4412
& \tfkaresult{0.3017}{0.0282} & \tfkaresult{0.3669}{0.0154} \\

& 720
& 0.4140 & 0.4210
& \tfkaresult{0.4072}{0.0072} & \tfkaresult{0.4301}{0.0057}
& 0.4498 & 0.4572
& \tfkaresult{0.4182}{0.0408} & \tfkaresult{0.4466}{0.0360}
& 0.4471 & 0.4402
& \tfkaresult{0.4541}{0.2349} & \tfkaresult{0.4528}{0.1040}
& 0.4646 & 0.4669
& \tfkaresult{0.3891}{0.0055} & \tfkaresult{0.4273}{0.0050} \\

\cmidrule{2-18}
& Avg
& 0.3666 & 0.3795
& \tfkaresult{0.3116}{0.0213} & \tfkaresult{0.3557}{0.0112}
& 0.4022 & 0.4094
& \tfkaresult{0.3643}{0.0223} & \tfkaresult{0.3977}{0.0183}
& 0.3884 & 0.3928
& \tfkaresult{0.3728}{0.0547} & \tfkaresult{0.3907}{0.0248}
& 0.4008 & 0.4291
& \tfkaresult{0.2984}{0.0225} & \tfkaresult{0.3600}{0.0127} \\

& $\Delta$
& -- & -- & \pctb{15.00\%} & \pctb{6.27\%}
& -- & -- & \pctb{9.42\%} & \pctb{2.86\%}
& -- & -- & \pctb{4.02\%} & \pctb{0.53\%}
& -- & -- & \pctb{25.55\%} & \pctb{16.10\%} \\
\midrule

\multirow{6}{*}{\rotatebox{90}{ETTm1}}
& 96
& 0.3319 & 0.3332
& \tfkaresult{0.2788}{0.0060} & \tfkaresult{0.3246}{0.0036}
& 0.3568 & 0.3668
& \tfkaresult{0.3061}{0.0091} & \tfkaresult{0.3454}{0.0029}
& 0.3786 & 0.3659
& \tfkaresult{0.2848}{0.0040} & \tfkaresult{0.3244}{0.0018}
& 0.7383 & 0.5409
& \tfkaresult{0.2969}{0.0005} & \tfkaresult{0.3441}{0.0004} \\

& 192
& 0.3870 & 0.3658
& \tfkaresult{0.3282}{0.0094} & \tfkaresult{0.3583}{0.0048}
& 0.4113 & 0.4024
& \tfkaresult{0.3653}{0.0248} & \tfkaresult{0.3824}{0.0117}
& 0.4667 & 0.4160
& \tfkaresult{0.3338}{0.0088} & \tfkaresult{0.3586}{0.0033}
& 0.7425 & 0.5526
& \tfkaresult{0.3404}{0.0023} & \tfkaresult{0.3712}{0.0014} \\

& 336
& 0.4363 & 0.3934
& \tfkaresult{0.3671}{0.0103} & \tfkaresult{0.3914}{0.0067}
& 0.4409 & 0.4261
& \tfkaresult{0.3788}{0.0302} & \tfkaresult{0.3978}{0.0179}
& 0.5558 & 0.4640
& \tfkaresult{0.3747}{0.0126} & \tfkaresult{0.3887}{0.0058}
& 0.7352 & 0.5578
& \tfkaresult{0.3716}{0.0029} & \tfkaresult{0.3928}{0.0016} \\

& 720
& 0.5279 & 0.4391
& \tfkaresult{0.4299}{0.0196} & \tfkaresult{0.4192}{0.0089}
& 0.5069 & 0.4685
& \tfkaresult{0.4779}{0.0698} & \tfkaresult{0.4581}{0.0420}
& 0.6541 & 0.5241
& \tfkaresult{0.4558}{0.0493} & \tfkaresult{0.4382}{0.0446}
& 0.7542 & 0.5725
& \tfkaresult{0.4284}{0.0030} & \tfkaresult{0.4206}{0.0018} \\

\cmidrule{2-18}
& Avg
& 0.4208 & 0.3829
& \tfkaresult{0.3510}{0.0109} & \tfkaresult{0.3734}{0.0057}
& 0.4290 & 0.4160
& \tfkaresult{0.3820}{0.0334} & \tfkaresult{0.3959}{0.0185}
& 0.5138 & 0.4425
& \tfkaresult{0.3623}{0.0102} & \tfkaresult{0.3775}{0.0114}
& 0.7426 & 0.5560
& \tfkaresult{0.3593}{0.0019} & \tfkaresult{0.3822}{0.0012} \\

& $\Delta$
& -- & -- & \pctb{16.59\%} & \pctb{2.48\%}
& -- & -- & \pctb{10.96\%} & \pctb{4.83\%}
& -- & -- & \pctb{29.49\%} & \pctb{14.69\%}
& -- & -- & \pctb{51.62\%} & \pctb{31.26\%} \\
\midrule

\multirow{6}{*}{\rotatebox{90}{ETTm2}}
& 96
& 0.1777 & 0.2448
& \tfkaresult{0.1664}{0.0022} & \tfkaresult{0.2368}{0.0015}
& 0.2046 & 0.2660
& \tfkaresult{0.1769}{0.0021} & \tfkaresult{0.2543}{0.0009}
& 0.1865 & 0.2501
& \tfkaresult{0.1656}{0.0045} & \tfkaresult{0.2392}{0.0072}
& 0.2260 & 0.3094
& \tfkaresult{0.1648}{0.0011} & \tfkaresult{0.2521}{0.0009} \\

& 192
& 0.2511 & 0.2935
& \tfkaresult{0.2322}{0.0031} & \tfkaresult{0.2808}{0.0021}
& 0.2927 & 0.3176
& \tfkaresult{0.2396}{0.0159} & \tfkaresult{0.2996}{0.0104}
& 0.2821 & 0.3100
& \tfkaresult{0.2251}{0.0401} & \tfkaresult{0.2832}{0.0085}
& 0.2979 & 0.3521
& \tfkaresult{0.2314}{0.0029} & \tfkaresult{0.2979}{0.0015} \\

& 336
& 0.3170 & 0.3359
& \tfkaresult{0.2903}{0.0044} & \tfkaresult{0.3204}{0.0029}
& 0.3644 & 0.3607
& \tfkaresult{0.2924}{0.0036} & \tfkaresult{0.3358}{0.0024}
& 0.4053 & 0.3745
& \tfkaresult{0.2844}{0.0174} & \tfkaresult{0.3258}{0.0098}
& 0.3434 & 0.3778
& \tfkaresult{0.2971}{0.0083} & \tfkaresult{0.3385}{0.0037} \\

& 720
& 0.4160 & 0.3950
& \tfkaresult{0.3741}{0.0065} & \tfkaresult{0.3757}{0.0026}
& 0.4644 & 0.4204
& \tfkaresult{0.4051}{0.0082} & \tfkaresult{0.4063}{0.0053}
& 0.5535 & 0.4501
& \tfkaresult{0.3800}{0.0180} & \tfkaresult{0.3918}{0.0093}
& 0.4302 & 0.4242
& \tfkaresult{0.3671}{0.0083} & \tfkaresult{0.3914}{0.0054} \\

\cmidrule{2-18}
& Avg
& 0.2905 & 0.3173
& \tfkaresult{0.2658}{0.0039} & \tfkaresult{0.3034}{0.0022}
& 0.3315 & 0.3412
& \tfkaresult{0.2785}{0.0039} & \tfkaresult{0.3240}{0.0022}
& 0.3569 & 0.3462
& \tfkaresult{0.2638}{0.0476} & \tfkaresult{0.3100}{0.0442}
& 0.3244 & 0.3659
& \tfkaresult{0.2651}{0.0041} & \tfkaresult{0.3200}{0.0021} \\

& $\Delta$
& -- & -- & \pctb{8.50\%} & \pctb{4.38\%}
& -- & -- & \pctb{15.99\%} & \pctb{5.04\%}
& -- & -- & \pctb{26.09\%} & \pctb{10.46\%}
& -- & -- & \pctb{18.28\%} & \pctb{12.54\%} \\
\midrule

\multirow{6}{*}{\rotatebox{90}{Exchange-rate}}
& 96
& 0.0992 & 0.2191
& \tfkaresult{0.0910}{0.0026} & \tfkaresult{0.2041}{0.0037}
& 0.1096 & 0.2331
& \tfkaresult{0.0905}{0.0012} & \tfkaresult{0.2140}{0.0006}
& 0.0903 & 0.2100
& \tfkaresult{0.0794}{0.0034} & \tfkaresult{0.1982}{0.0032}
& 0.1264 & 0.2525
& \tfkaresult{0.1042}{0.0065} & \tfkaresult{0.2214}{0.0053} \\

& 192
& 0.1944 & 0.3140
& \tfkaresult{0.1886}{0.0150} & \tfkaresult{0.2989}{0.0110}
& 0.1977 & 0.3151
& \tfkaresult{0.1866}{0.0260} & \tfkaresult{0.3080}{0.0251}
& 0.1888 & 0.3084
& \tfkaresult{0.1575}{0.0177} & \tfkaresult{0.2878}{0.0147}
& 0.2543 & 0.3646
& \tfkaresult{0.2166}{0.0121} & \tfkaresult{0.3303}{0.0080} \\

& 336
& 0.3420 & 0.4238
& \tfkaresult{0.3408}{0.0310} & \tfkaresult{0.4180}{0.0189}
& 0.3625 & 0.4333
& \tfkaresult{0.3582}{0.0946} & \tfkaresult{0.4318}{0.0745}
& 0.3459 & 0.4307
& \tfkaresult{0.2609}{0.0451} & \tfkaresult{0.3804}{0.0328}
& 0.4430 & 0.4916
& \tfkaresult{0.3893}{0.0096} & \tfkaresult{0.4536}{0.0050} \\

& 720
& 0.8493 & 0.6915
& \tfkaresult{0.7826}{0.1265} & \tfkaresult{0.6717}{0.0472}
& 0.9421 & 0.7260
& \tfkaresult{0.8317}{0.2839} & \tfkaresult{0.6871}{0.1766}
& 0.8930 & 0.6939
& \tfkaresult{0.5090}{0.1752} & \tfkaresult{0.5536}{0.0888}
& 1.0197 & 0.7622
& \tfkaresult{1.0017}{0.1151} & \tfkaresult{0.7355}{0.0425} \\

\cmidrule{2-18}
& Avg
& 0.3712 & 0.4121
& \tfkaresult{0.3508}{0.0435} & \tfkaresult{0.3982}{0.0200}
& 0.4030 & 0.4269
& \tfkaresult{0.3668}{0.1008} & \tfkaresult{0.4102}{0.0691}
& 0.3795 & 0.4108
& \tfkaresult{0.2517}{0.0573} & \tfkaresult{0.3550}{0.0333}
& 0.4609 & 0.4677
& \tfkaresult{0.4280}{0.0276} & \tfkaresult{0.4352}{0.0107} \\

& $\Delta$
& -- & -- & \pctb{5.50\%} & \pctb{3.37\%}
& -- & -- & \pctb{8.98\%} & \pctb{3.91\%}
& -- & -- & \pctb{33.68\%} & \pctb{13.58\%}
& -- & -- & \pctb{7.14\%} & \pctb{6.95\%} \\
\midrule

\multirow{6}{*}{\rotatebox{90}{Electricity}}
& 96
& 0.1242 & 0.2105
& \tfkaresult{0.1143}{0.0036} & \tfkaresult{0.2068}{0.0039}
& -- & -- & -- & --
& 0.1282 & 0.2146
& \tfkaresult{0.1263}{0.0028} & \tfkaresult{0.2134}{0.0076}
& 0.3964 & 0.4909
& \tfkaresult{0.1274}{0.0044} & \tfkaresult{0.2337}{0.0002} \\

& 192
& 0.1428 & 0.2287
& \tfkaresult{0.1361}{0.0054} & \tfkaresult{0.2263}{0.0052}
& -- & -- & -- & --
& 0.1473 & 0.2325
& \tfkaresult{0.1447}{0.0064} & \tfkaresult{0.2310}{0.0090}
& 0.4401 & 0.5191
& \tfkaresult{0.1463}{0.0045} & \tfkaresult{0.2502}{0.0013} \\

& 336
& 0.1613 & 0.2482
& \tfkaresult{0.1576}{0.0075} & \tfkaresult{0.2464}{0.0067}
& -- & -- & -- & --
& 0.1662 & 0.2517
& \tfkaresult{0.1614}{0.0116} & \tfkaresult{0.2483}{0.0115}
& 0.4882 & 0.5495
& \tfkaresult{0.1671}{0.0035} & \tfkaresult{0.2687}{0.0019} \\

& 720
& 0.2069 & 0.2882
& \tfkaresult{0.2033}{0.0112} & \tfkaresult{0.2832}{0.0089}
& -- & -- & -- & --
& 0.2117 & 0.2913
& \tfkaresult{0.1976}{0.0214} & \tfkaresult{0.2808}{0.0169}
& 0.6321 & 0.6272
& \tfkaresult{0.2144}{0.0010} & \tfkaresult{0.3067}{0.0009} \\

\cmidrule{2-18}
& Avg
& 0.1588 & 0.2439
& \tfkaresult{0.1528}{0.0065} & \tfkaresult{0.2407}{0.0059}
& -- & -- & -- & --
& 0.1634 & 0.2475
& \tfkaresult{0.1575}{0.0105} & \tfkaresult{0.2434}{0.0113}
& 0.4892 & 0.5467
& \tfkaresult{0.1638}{0.0025} & \tfkaresult{0.2648}{0.0006} \\

& $\Delta$
& -- & -- & \pctb{3.78\%} & \pctb{1.31\%}
& -- & -- & -- & --
& -- & -- & \pctb{3.61\%} & \pctb{1.66\%}
& -- & -- & \pctb{66.52\%} & \pctb{51.56\%} \\
\midrule

\multirow{6}{*}{\rotatebox{90}{Weather}}
& 96
& 0.1775 & 0.2102
& \tfkaresult{0.1488}{0.0007} & \tfkaresult{0.1822}{0.0009}
& -- & -- & -- & --
& 0.1742 & 0.2015
& \tfkaresult{0.1542}{0.0071} & \tfkaresult{0.1882}{0.0125}
& 0.1221 & 0.1565
& \tfkaresult{0.1169}{0.0112} & \tfkaresult{0.1537}{0.0153} \\

& 192
& 0.2242 & 0.2535
& \tfkaresult{0.1948}{0.0015} & \tfkaresult{0.2286}{0.0013}
& -- & -- & -- & --
& 0.2239 & 0.2476
& \tfkaresult{0.2069}{0.0082} & \tfkaresult{0.2459}{0.0120}
& 0.1671 & 0.1996
& \tfkaresult{0.1585}{0.0146} & \tfkaresult{0.1959}{0.0177} \\

& 336
& 0.2838 & 0.2947
& \tfkaresult{0.2479}{0.0019} & \tfkaresult{0.2702}{0.0017}
& -- & -- & -- & --
& 0.2949 & 0.2961
& \tfkaresult{0.2586}{0.0195} & \tfkaresult{0.2855}{0.0184}
& 0.2358 & 0.2540
& \tfkaresult{0.2189}{0.0112} & \tfkaresult{0.2481}{0.0133} \\

& 720
& 0.3670 & 0.3465
& \tfkaresult{0.3213}{0.0043} & \tfkaresult{0.3207}{0.0030}
& -- & -- & -- & --
& 0.3989 & 0.3558
& \tfkaresult{0.3309}{0.0610} & \tfkaresult{0.3360}{0.0427}
& 0.3610 & 0.3391
& \tfkaresult{0.3245}{0.0038} & \tfkaresult{0.3241}{0.0058} \\

\cmidrule{2-18}
& Avg
& 0.2631 & 0.2762
& \tfkaresult{0.2282}{0.0020} & \tfkaresult{0.2504}{0.0016}
& -- & -- & -- & --
& 0.2730 & 0.2753
& \tfkaresult{0.2377}{0.0103} & \tfkaresult{0.2639}{0.0080}
& 0.2215 & 0.2373
& \tfkaresult{0.2047}{0.0099} & \tfkaresult{0.2304}{0.0129} \\

& $\Delta$
& -- & -- & \pctb{13.26\%} & \pctb{9.34\%}
& -- & -- & -- & --
& -- & -- & \pctb{12.93\%} & \pctb{4.14\%}
& -- & -- & \pctb{7.58\%} & \pctb{2.91\%} \\
\midrule

\multirow{6}{*}{\rotatebox{90}{Traffic}}
& 96
& 0.3870 & 0.2409
& \tfkaresult{0.3407}{0.0038} & \tfkaresult{0.2129}{0.0015}
& -- & -- & -- & --
& 0.3546 & 0.2248
& \tfkaresult{0.3498}{0.0060} & \tfkaresult{0.2214}{0.0007}
& 0.3273 & 0.2204
& \tfkaresult{0.3257}{0.0025} & \tfkaresult{0.2181}{0.0027} \\

& 192
& 0.4137 & 0.2531
& \tfkaresult{0.3592}{0.0035} & \tfkaresult{0.2224}{0.0056}
& -- & -- & -- & --
& 0.3789 & 0.2355
& \tfkaresult{0.3728}{0.0054} & \tfkaresult{0.2318}{0.0097}
& 0.3538 & 0.2346
& \tfkaresult{0.3511}{0.0025} & \tfkaresult{0.2314}{0.0028} \\

& 336
& 0.4355 & 0.2633
& \tfkaresult{0.3761}{0.0036} & \tfkaresult{0.2332}{0.0050}
& -- & -- & -- & --
& 0.3959 & 0.2439
& \tfkaresult{0.3878}{0.0057} & \tfkaresult{0.2393}{0.0084}
& 0.3779 & 0.2477
& \tfkaresult{0.3758}{0.0023} & \tfkaresult{0.2451}{0.0028} \\

& 720
& 0.5033 & 0.2934
& \tfkaresult{0.4262}{0.0033} & \tfkaresult{0.2617}{0.0037}
& -- & -- & -- & --
& 0.4337 & 0.2665
& \tfkaresult{0.4231}{0.0052} & \tfkaresult{0.2596}{0.0059}
& 0.4196 & 0.2722
& \tfkaresult{0.4170}{0.0022} & \tfkaresult{0.2694}{0.0024} \\

\cmidrule{2-18}
& Avg
& 0.4349 & 0.2627
& \tfkaresult{0.3756}{0.0063} & \tfkaresult{0.2326}{0.0044}
& -- & -- & -- & --
& 0.3908 & 0.2427
& \tfkaresult{0.3834}{0.0056} & \tfkaresult{0.2380}{0.0007}
& 0.3696 & 0.2437
& \tfkaresult{0.3674}{0.0085} & \tfkaresult{0.2410}{0.0091} \\

& $\Delta$
& -- & -- & \pctb{13.64\%} & \pctb{11.46\%}
& -- & -- & -- & --
& -- & -- & \pctb{1.89\%} & \pctb{1.94\%}
& -- & -- & \pctb{0.60\%} & \pctb{1.11\%} \\

\bottomrule
\end{tabular}%
}
\end{small}
\end{table*}
\endgroup

%% file: table/app/chrons_bolt_baseline.tex
\begin{table*}[!p]
\centering
\caption{Detailed forecasting results on ChronosBolt-base.
Avg denotes the arithmetic mean over forecasting horizons
$H \in \{96,192,336,720\}$.
For \method, $\pm$ denotes the standard deviation over five independent runs, where reported.
The best and second-best results are highlighted in \firstres{purple}
and underlined in \secondres{green}, respectively;
ties are determined at the displayed precision.
Relative error reductions \pctb{percentages} and increases \pctr{percentages}
are computed from the displayed Avg values against zero-shot forecasting.
OOM: out of memory.}
\label{tab:chronosbolt_main_appendix}

\begingroup
\small
\setlength{\tabcolsep}{3.5pt}
\renewcommand{\arraystretch}{1.0}
\newcommand{\tfkastd}[1]{\textcolor{gray}{\scriptsize$\pm#1$}}

\resizebox{\textwidth}{!}{%
\begin{tabular}{c|c|cc|cc|cc|cc|cc|cc|cc}
\toprule
\multirow{2}{*}{Dataset} &
\multirow{2}{*}{$H$} &
\multicolumn{2}{c|}{Zero-shot} &
\multicolumn{2}{c|}{RAFT} &
\multicolumn{2}{c|}{TS-RAG} &
\multicolumn{2}{c|}{TS-Memory} &
\multicolumn{2}{c|}{SARAF} &
\multicolumn{2}{c|}{MEMTS} &
\multicolumn{2}{c}{\method} \\
& &
MSE & MAE &
MSE & MAE &
MSE & MAE &
MSE & MAE &
MSE & MAE &
MSE & MAE &
MSE & MAE \\
\midrule

\multirow{6}{*}{\rotatebox{90}{ETTh1}}
& 96
& 0.3849 & 0.3798
& 0.3782 & 0.3798
& 0.3780 & 0.3787
& 0.3801 & 0.3807
& \secondres{0.3758} & 0.4034
& 0.3768 & \secondres{0.3732}
& \firstres{0.3570}\tfkastd{0.0087}
& \firstres{0.3720}\tfkastd{0.0047} \\
& 192
& 0.4417 & 0.4122
& 0.4308 & 0.4106
& 0.4319 & 0.4114
& 0.4237 & 0.4131
& \firstres{0.3948} & 0.4126
& 0.4271 & \secondres{0.3991}
& \secondres{0.3978}\tfkastd{0.0112}
& \firstres{0.3964}\tfkastd{0.0064} \\
& 336
& 0.4814 & 0.4331
& 0.4648 & 0.4299
& 0.4699 & 0.4337
& 0.4645 & 0.4301
& \secondres{0.4223} & 0.4313
& 0.4511 & \secondres{0.4171}
& \firstres{0.4208}\tfkastd{0.0128}
& \firstres{0.4131}\tfkastd{0.0058} \\
& 720
& 0.4836 & 0.4489
& 0.4734 & 0.4501
& 0.4893 & 0.4619
& 0.4557 & 0.4556
& \firstres{0.4452} & 0.4631
& 0.4778 & \secondres{0.4479}
& \secondres{0.4456}\tfkastd{0.0102}
& \firstres{0.4408}\tfkastd{0.0037} \\
\cmidrule{2-16}
& Avg
& 0.4479 & 0.4185
& 0.4368 & 0.4176
& 0.4423 & 0.4214
& 0.4310 & 0.4199
& \secondres{0.4095} & 0.4276
& 0.4332 & \secondres{0.4093}
& \firstres{0.4053}\tfkastd{0.0105}
& \firstres{0.4056}\tfkastd{0.0049} \\
& $\Delta$
& -- & --
& \pctb{2.48\%} & \pctb{0.22\%}
& \pctb{1.25\%} & \pctr{0.69\%}
& \pctb{3.77\%} & \pctr{0.33\%}
& \pctb{8.57\%} & \pctr{2.17\%}
& \pctb{3.28\%} & \pctb{2.20\%}
& \pctb{9.51\%} & \pctb{3.08\%} \\

\midrule
\multirow{6}{*}{\rotatebox{90}{ETTh2}}
& 96
& 0.2895 & 0.3251
& 0.2856 & 0.3245
& 0.2834 & 0.3258
& 0.2865 & 0.3250
& 0.2709 & 0.3353
& \secondres{0.2534} & \secondres{0.3040}
& \firstres{0.2424}\tfkastd{0.0197}
& \firstres{0.2982}\tfkastd{0.0124} \\
& 192
& 0.3598 & 0.3695
& 0.3521 & 0.3662
& 0.3514 & 0.3698
& 0.3557 & 0.3688
& 0.3300 & 0.3765
& \secondres{0.2982} & \secondres{0.3356}
& \firstres{0.2812}\tfkastd{0.0309}
& \firstres{0.3310}\tfkastd{0.0170} \\
& 336
& 0.4029 & 0.4025
& 0.3898 & 0.3965
& 0.3934 & 0.4024
& 0.3975 & 0.4011
& 0.3536 & 0.4028
& \secondres{0.3343} & \secondres{0.3659}
& \firstres{0.3157}\tfkastd{0.0306}
& \firstres{0.3633}\tfkastd{0.0155} \\
& 720
& 0.4140 & \secondres{0.4210}
& \firstres{0.4020} & 0.4233
& 0.4183 & 0.4432
& 0.4140 & \secondres{0.4210}
& 0.4100 & 0.4524
& \secondres{0.4035} & \firstres{0.4180}
& 0.4072\tfkastd{0.0072}
& 0.4301\tfkastd{0.0057} \\
\cmidrule{2-16}
& Avg
& 0.3666 & 0.3795
& 0.3574 & 0.3776
& 0.3616 & 0.3853
& 0.3634 & 0.3790
& 0.3411 & 0.3918
& \secondres{0.3224} & \secondres{0.3559}
& \firstres{0.3116}\tfkastd{0.0213}
& \firstres{0.3557}\tfkastd{0.0112} \\
& $\Delta$
& -- & --
& \pctb{2.51\%} & \pctb{0.50\%}
& \pctb{1.36\%} & \pctr{1.53\%}
& \pctb{0.87\%} & \pctb{0.13\%}
& \pctb{6.96\%} & \pctr{3.24\%}
& \pctb{12.06\%} & \pctb{6.22\%}
& \pctb{15.00\%} & \pctb{6.27\%} \\

\midrule
\multirow{6}{*}{\rotatebox{90}{ETTm1}}
& 96
& 0.3319 & 0.3332
& 0.3134 & 0.3274
& 0.3237 & 0.3338
& \secondres{0.2980} & \secondres{0.3269}
& 0.3071 & 0.3539
& 0.3245 & 0.3387
& \firstres{0.2788}\tfkastd{0.0060}
& \firstres{0.3246}\tfkastd{0.0036} \\
& 192
& 0.3870 & 0.3658
& 0.3688 & 0.3604
& 0.3755 & 0.3648
& 0.3505 & \secondres{0.3586}
& \secondres{0.3332} & 0.3686
& 0.3727 & 0.3681
& \firstres{0.3282}\tfkastd{0.0094}
& \firstres{0.3583}\tfkastd{0.0048} \\
& 336
& 0.4363 & 0.3934
& 0.4149 & \secondres{0.3878}
& 0.4211 & 0.3910
& 0.3984 & \firstres{0.3826}
& \firstres{0.3638} & 0.3883
& 0.4279 & 0.3958
& \secondres{0.3671}\tfkastd{0.0103}
& 0.3914\tfkastd{0.0067} \\
& 720
& 0.5279 & 0.4391
& 0.4928 & 0.4315
& 0.5028 & 0.4357
& 0.4556 & 0.4206
& \firstres{0.4095} & \firstres{0.4157}
& 0.5133 & 0.4371
& \secondres{0.4299}\tfkastd{0.0196}
& \secondres{0.4192}\tfkastd{0.0089} \\
\cmidrule{2-16}
& Avg
& 0.4208 & 0.3829
& 0.3975 & 0.3768
& 0.4058 & 0.3813
& 0.3756 & \firstres{0.3722}
& \secondres{0.3534} & 0.3816
& 0.4096 & 0.3849
& \firstres{0.3510}\tfkastd{0.0109}
& \secondres{0.3734}\tfkastd{0.0057} \\
& $\Delta$
& -- & --
& \pctb{5.54\%} & \pctb{1.59\%}
& \pctb{3.56\%} & \pctb{0.42\%}
& \pctb{10.74\%} & \pctb{2.79\%}
& \pctb{16.02\%} & \pctb{0.34\%}
& \pctb{2.66\%} & \pctr{0.52\%}
& \pctb{16.59\%} & \pctb{2.48\%} \\

\midrule
\multirow{6}{*}{\rotatebox{90}{ETTm2}}
& 96
& 0.1777 & 0.2448
& 0.1741 & \secondres{0.2430}
& 0.1785 & 0.2486
& 0.1681 & 0.2440
& \firstres{0.1632} & 0.2544
& 0.1816 & 0.2528
& \secondres{0.1664}\tfkastd{0.0022}
& \firstres{0.2368}\tfkastd{0.0015} \\
& 192
& 0.2511 & 0.2935
& 0.2431 & \secondres{0.2896}
& 0.2438 & 0.2935
& 0.2334 & 0.2897
& \firstres{0.2180} & 0.2931
& 0.2552 & 0.3014
& \secondres{0.2322}\tfkastd{0.0031}
& \firstres{0.2808}\tfkastd{0.0021} \\
& 336
& 0.3170 & 0.3359
& 0.3056 & 0.3315
& 0.3041 & 0.3341
& 0.3039 & 0.3346
& \firstres{0.2695} & \secondres{0.3284}
& 0.3234 & 0.3439
& \secondres{0.2903}\tfkastd{0.0044}
& \firstres{0.3204}\tfkastd{0.0029} \\
& 720
& 0.4160 & 0.3950
& 0.4011 & 0.3939
& 0.4051 & 0.3962
& 0.3948 & 0.3970
& \firstres{0.3573} & \secondres{0.3851}
& 0.4196 & 0.4007
& \secondres{0.3741}\tfkastd{0.0065}
& \firstres{0.3757}\tfkastd{0.0026} \\
\cmidrule{2-16}
& Avg
& 0.2905 & 0.3173
& 0.2810 & \secondres{0.3145}
& 0.2829 & 0.3181
& 0.2751 & 0.3163
& \firstres{0.2520} & 0.3153
& 0.2950 & 0.3247
& \secondres{0.2658}\tfkastd{0.0039}
& \firstres{0.3034}\tfkastd{0.0022} \\
& $\Delta$
& -- & --
& \pctb{3.27\%} & \pctb{0.88\%}
& \pctb{2.62\%} & \pctr{0.25\%}
& \pctb{5.30\%} & \pctb{0.32\%}
& \pctb{13.25\%} & \pctb{0.63\%}
& \pctr{1.55\%} & \pctr{2.33\%}
& \pctb{8.50\%} & \pctb{4.38\%} \\

\midrule
\multirow{6}{*}{\rotatebox{90}{Exchange-rate}}
& 96
& 0.0992 & 0.2191
& 0.0960 & 0.2155
& 0.0933 & 0.2131
& 0.0983 & 0.2197
& \firstres{0.0881} & \secondres{0.2059}
& 0.1009 & 0.2160
& \secondres{0.0910}\tfkastd{0.0026}
& \firstres{0.2041}\tfkastd{0.0037} \\
& 192
& 0.1944 & 0.3140
& 0.1887 & 0.3106
& 0.1895 & 0.3117
& \firstres{0.1872} & \secondres{0.3096}
& 0.1952 & 0.3142
& 0.2050 & 0.3136
& \secondres{0.1886}\tfkastd{0.0150}
& \firstres{0.2989}\tfkastd{0.0110} \\
& 336
& 0.3420 & 0.4238
& \firstres{0.3335} & 0.4184
& \secondres{0.3364} & \secondres{0.4182}
& 0.3825 & 0.4604
& 0.3611 & 0.4310
& 0.3643 & 0.4297
& 0.3408\tfkastd{0.0310}
& \firstres{0.4180}\tfkastd{0.0189} \\
& 720
& 0.8493 & 0.6915
& 0.8564 & 0.6893
& 0.8483 & 0.6887
& 0.9240 & 0.7359
& 0.9335 & 0.7256
& \secondres{0.8399} & \secondres{0.6870}
& \firstres{0.7826}\tfkastd{0.1265}
& \firstres{0.6717}\tfkastd{0.0472} \\
\cmidrule{2-16}
& Avg
& 0.3712 & 0.4121
& 0.3687 & 0.4085
& \secondres{0.3669} & \secondres{0.4079}
& 0.3980 & 0.4314
& 0.3945 & 0.4192
& 0.3775 & 0.4116
& \firstres{0.3508}\tfkastd{0.0435}
& \firstres{0.3982}\tfkastd{0.0200} \\
& $\Delta$
& -- & --
& \pctb{0.67\%} & \pctb{0.87\%}
& \pctb{1.16\%} & \pctb{1.02\%}
& \pctr{7.22\%} & \pctr{4.68\%}
& \pctr{6.28\%} & \pctr{1.72\%}
& \pctr{1.70\%} & \pctb{0.12\%}
& \pctb{5.50\%} & \pctb{3.37\%} \\

\midrule
\multirow{6}{*}{\rotatebox{90}{Weather}}
& 96
& 0.1775 & 0.2102
& 0.1671 & \secondres{0.2022}
& 0.1760 & 0.2148
& \secondres{0.1636} & 0.2157
& 0.1694 & 0.2315
& 0.1706 & 0.2333
& \firstres{0.1488}\tfkastd{0.0007}
& \firstres{0.1822}\tfkastd{0.0009} \\
& 192
& 0.2242 & 0.2535
& 0.2138 & \secondres{0.2462}
& 0.2220 & 0.2572
& 0.2152 & 0.2626
& 0.2156 & 0.2725
& \secondres{0.2131} & 0.2656
& \firstres{0.1948}\tfkastd{0.0015}
& \firstres{0.2286}\tfkastd{0.0013} \\
& 336
& 0.2838 & 0.2947
& 0.2763 & \secondres{0.2867}
& 0.2839 & 0.2981
& 0.2638 & 0.3005
& \secondres{0.2630} & 0.3091
& 0.2793 & 0.3061
& \firstres{0.2479}\tfkastd{0.0019}
& \firstres{0.2702}\tfkastd{0.0017} \\
& 720
& 0.3670 & 0.3465
& 0.3585 & \secondres{0.3367}
& 0.3639 & 0.3474
& 0.3357 & 0.3508
& \secondres{0.3336} & 0.3625
& 0.3669 & 0.3606
& \firstres{0.3213}\tfkastd{0.0043}
& \firstres{0.3207}\tfkastd{0.0030} \\
\cmidrule{2-16}
& Avg
& 0.2631 & 0.2762
& 0.2539 & \secondres{0.2680}
& 0.2615 & 0.2794
& \secondres{0.2446} & 0.2824
& 0.2454 & 0.2939
& 0.2575 & 0.2914
& \firstres{0.2282}\tfkastd{0.0020}
& \firstres{0.2504}\tfkastd{0.0016} \\
& $\Delta$
& -- & --
& \pctb{3.50\%} & \pctb{2.97\%}
& \pctb{0.61\%} & \pctr{1.16\%}
& \pctb{7.03\%} & \pctr{2.24\%}
& \pctb{6.73\%} & \pctr{6.41\%}
& \pctb{2.13\%} & \pctr{5.50\%}
& \pctb{13.26\%} & \pctb{9.34\%} \\

\midrule
\multirow{6}{*}{\rotatebox{90}{Electricity}}
& 96
& 0.1242 & 0.2105
& 0.1226 & 0.2104
& \secondres{0.1224} & \secondres{0.2103}
& 0.1235 & 0.2107
& 0.1359 & 0.2370
& OOM & OOM
& \firstres{0.1143}\tfkastd{0.0036}
& \firstres{0.2068}\tfkastd{0.0039} \\
& 192
& 0.1428 & 0.2287
& \secondres{0.1410} & \secondres{0.2286}
& 0.1411 & \secondres{0.2286}
& 0.1419 & \secondres{0.2286}
& 0.1470 & 0.2464
& OOM & OOM
& \firstres{0.1361}\tfkastd{0.0054}
& \firstres{0.2263}\tfkastd{0.0052} \\
& 336
& 0.1613 & 0.2482
& 0.1587 & 0.2482
& 0.1590 & 0.2483
& 0.1599 & \secondres{0.2478}
& \firstres{0.1563} & 0.2578
& OOM & OOM
& \secondres{0.1576}\tfkastd{0.0075}
& \firstres{0.2464}\tfkastd{0.0067} \\
& 720
& 0.2069 & 0.2882
& \secondres{0.2032} & 0.2900
& 0.2042 & 0.2907
& 0.2037 & 0.2863
& \firstres{0.1725} & \firstres{0.2738}
& OOM & OOM
& 0.2033\tfkastd{0.0112}
& \secondres{0.2832}\tfkastd{0.0089} \\
\cmidrule{2-16}
& Avg
& 0.1588 & 0.2439
& 0.1564 & 0.2443
& 0.1567 & 0.2445
& 0.1573 & \secondres{0.2434}
& \secondres{0.1529} & 0.2538
& OOM & OOM
& \firstres{0.1528}\tfkastd{0.0065}
& \firstres{0.2407}\tfkastd{0.0059} \\
& $\Delta$
& -- & --
& \pctb{1.51\%} & \pctr{0.16\%}
& \pctb{1.32\%} & \pctr{0.25\%}
& \pctb{0.94\%} & \pctb{0.21\%}
& \pctb{3.72\%} & \pctr{4.06\%}
& -- & --
& \pctb{3.78\%} & \pctb{1.31\%} \\

\midrule
\multirow{6}{*}{\rotatebox{90}{Traffic}}
& 96
& 0.3870 & 0.2409
& 0.3835 & 0.2402
& 0.3833 & \secondres{0.2401}
& \secondres{0.3788} & \secondres{0.2401}
& 0.3979 & 0.2924
& OOM & OOM
& \firstres{0.3407}\tfkastd{0.0038}
& \firstres{0.2129}\tfkastd{0.0015} \\
& 192
& 0.4137 & 0.2531
& 0.4095 & 0.2522
& 0.4097 & 0.2522
& 0.4086 & \secondres{0.2520}
& \secondres{0.4052} & 0.2914
& OOM & OOM
& \firstres{0.3592}\tfkastd{0.0035}
& \firstres{0.2224}\tfkastd{0.0056} \\
& 336
& 0.4355 & 0.2633
& 0.4286 & \secondres{0.2618}
& 0.4288 & 0.2619
& 0.4316 & 0.2624
& \secondres{0.4109} & 0.2915
& OOM & OOM
& \firstres{0.3761}\tfkastd{0.0036}
& \firstres{0.2332}\tfkastd{0.0050} \\
& 720
& 0.5033 & 0.2934
& 0.4843 & 0.2888
& 0.4850 & 0.2894
& 0.4902 & \secondres{0.2886}
& \secondres{0.4369} & 0.3019
& OOM & OOM
& \firstres{0.4262}\tfkastd{0.0033}
& \firstres{0.2617}\tfkastd{0.0037} \\
\cmidrule{2-16}
& Avg
& 0.4349 & 0.2627
& 0.4265 & \secondres{0.2608}
& 0.4267 & 0.2609
& 0.4273 & \secondres{0.2608}
& \secondres{0.4127} & 0.2943
& OOM & OOM
& \firstres{0.3756} \textcolor{gray}{\text{\scriptsize±0.0063}}
& \firstres{0.2326} \textcolor{gray}{\text{\scriptsize±0.0044}}\\
& $\Delta$
& -- & --
& \pctb{1.93\%} & \pctb{0.72\%}
& \pctb{1.89\%} & \pctb{0.69\%}
& \pctb{1.75\%} & \pctb{0.72\%}
& \pctb{5.10\%} & \pctr{12.03\%}
& -- & --
& \pctb{13.64\%} & \pctb{11.46\%} \\

\bottomrule
\end{tabular}%
}
\endgroup
\end{table*}

%% file: reference.bib
@article{qin2025cora,
  title={Cora: Covariate-aware adaptation of time series foundation models},
  author={Qin, Guo and Chen, Zhi and Liu, Yong and Shi, Zhiyuan and Liu, Haixuan and Huang, Xiangdong and Wang, Jianmin and Long, Mingsheng},
  journal={arXiv preprint arXiv:2510.12681},
  year={2025}
}

@article{chen2025select,
  title={Select, then balance: A plug-and-play framework for exogenous-aware spatio-temporal forecasting},
  author={Chen, Wei and Wu, Yuqian and Zhu, Yuanshao and Hao, Xixuan and Wang, Shiyu and Liang, Yuxuan},
  journal={arXiv e-prints},
  pages={arXiv--2509},
  year={2025}
}

@misc{ansari2024chronos,
  title={Chronos: Learning the language of time series},
  author={Ansari, Abdul Fatir and Stella, Lorenzo and Turkmen, Caner and Zhang, Xiyuan and Mercado, Pedro and Shen, Huibin and Shchur, Oleksandr and Rangapuram, Syama Sundar and Arango, Sebastian Pineda and Kapoor, Shubham and others},
  journal={arXiv preprint arXiv:2403.07815},
  year={2024}
}

@misc{das2024decoder,
  title={A decoder-only foundation model for time-series forecasting},
  author={Das, Abhimanyu and Kong, Weihao and Sen, Rajat and Zhou, Yichen},
  booktitle={Forty-first International Conference on Machine Learning},
  year={2024}
}

@misc{ansari2025chronos,
  title={Chronos-2: From univariate to universal forecasting},
  author={Ansari, Abdul Fatir and Shchur, Oleksandr and K{\"u}ken, Jaris and Auer, Andreas and Han, Boran and Mercado, Pedro and Rangapuram, Syama Sundar and Shen, Huibin and Stella, Lorenzo and Zhang, Xiyuan and others},
  journal={arXiv preprint arXiv:2510.15821},
  year={2025}
}

@article{lyu2026ts,
  title={Ts-memory: Plug-and-play memory for time series foundation models},
  author={Lyu, Sisuo and Zhong, Siru and Chen, Tiegang and Ruan, Weilin and Liu, Qingxiang and Lv, Taiqiang and Wen, Qingsong and Wong, Raymond Chi-Wing and Liang, Yuxuan},
  journal={arXiv preprint arXiv:2602.11550},
  year={2026}
}

@article{ye2026empowering,
  title={Empowering time series analysis with foundation models: A comprehensive survey},
  author={Ye, Jiexia and Yu, Yongzi and Zhang, Weiqi and Wang, Le and Li, Jia and Tsung, Fugee},
  journal={Information Fusion},
  pages={104601},
  year={2026},
  publisher={Elsevier}
}

@inproceedings{chen2026urbanfm,
  title={UrbanFM: Scaling urban spatio-temporal foundation models},
  author={Chen, Wei and Wu, Yuqian and Chen, Junle and Zhou, Xiaofang and Liang, Yuxuan},
  booktitle={Proceedings of the 32nd ACM SIGKDD Conference on Knowledge Discovery and Data Mining V. 2},
  pages={10635--10645},
  year={2026}
}

@misc{rasul2024lagllama,
      title={Lag-Llama: Towards Foundation Models for Probabilistic Time Series Forecasting}, 
      author={Kashif Rasul and Arjun Ashok and Andrew Robert Williams and Hena Ghonia and Rishika Bhagwatkar and Arian Khorasani and Mohammad Javad Darvishi Bayazi and George Adamopoulos and Roland Riachi and Nadhir Hassen and Marin Biloš and Sahil Garg and Anderson Schneider and Nicolas Chapados and Alexandre Drouin and Valentina Zantedeschi and Yuriy Nevmyvaka and Irina Rish},
      year={2024},
      eprint={2310.08278},
      archivePrefix={arXiv},
      primaryClass={cs.LG},
      url={https://arxiv.org/abs/2310.08278}, 
}

@misc{garza2024timegpt1,
      title={TimeGPT-1}, 
      author={Azul Garza and Cristian Challu and Max Mergenthaler-Canseco},
      year={2024},
      eprint={2310.03589},
      archivePrefix={arXiv},
      primaryClass={cs.LG},
      url={https://arxiv.org/abs/2310.03589}, 
}

@misc{gao2024units,
      title={UniTS: A Unified Multi-Task Time Series Model}, 
      author={Shanghua Gao and Teddy Koker and Owen Queen and Thomas Hartvigsen and Theodoros Tsiligkaridis and Marinka Zitnik},
      year={2024},
      eprint={2403.00131},
      archivePrefix={arXiv},
      primaryClass={cs.LG},
      url={https://arxiv.org/abs/2403.00131}, 
}

@misc{woo2024unified,
      title={Unified Training of Universal Time Series Forecasting Transformers}, 
      author={Gerald Woo and Chenghao Liu and Akshat Kumar and Caiming Xiong and Silvio Savarese and Doyen Sahoo},
      year={2024},
      eprint={2402.02592},
      archivePrefix={arXiv},
      primaryClass={cs.LG},
      url={https://arxiv.org/abs/2402.02592}, 
}

@misc{goswami2024moment,
      title={MOMENT: A Family of Open Time-series Foundation Models}, 
      author={Mononito Goswami and Konrad Szafer and Arjun Choudhry and Yifu Cai and Shuo Li and Artur Dubrawski},
      year={2024},
      eprint={2402.03885},
      archivePrefix={arXiv},
      primaryClass={cs.LG},
      url={https://arxiv.org/abs/2402.03885}, 
}

@misc{liu2024timer,
      title={Timer: Generative Pre-trained Transformers Are Large Time Series Models}, 
      author={Yong Liu and Haoran Zhang and Chenyu Li and Xiangdong Huang and Jianmin Wang and Mingsheng Long},
      year={2024},
      eprint={2402.02368},
      archivePrefix={arXiv},
      primaryClass={cs.LG},
      url={https://arxiv.org/abs/2402.02368}, 
}

@misc{liu2025timerx,
      title={Timer-XL: Long-Context Transformers for Unified Time Series Forecasting}, 
      author={Yong Liu and Guo Qin and Xiangdong Huang and Jianmin Wang and Mingsheng Long},
      year={2025},
      eprint={2410.04803},
      archivePrefix={arXiv},
      primaryClass={cs.LG},
      url={https://arxiv.org/abs/2410.04803}, 
}

@article{ekambaram2024tiny,
  title={Tiny time mixers (ttms): Fast pre-trained models for enhanced zero/few-shot forecasting of multivariate time series},
  author={Ekambaram, Vijay and Jati, Arindam and Dayama, Pankaj and Mukherjee, Sumanta and Nguyen, Nam H and Gifford, Wesley M and Reddy, Chandra and Kalagnanam, Jayant},
  journal={Advances in Neural Information Processing Systems},
  volume={37},
  pages={74147--74181},
  year={2024}
}

@article{tire2024retrieval,
  title={Retrieval augmented time series forecasting},
  author={Tire, Kutay and Taga, Ege Onur and Ildiz, Muhammed Emrullah and Oymak, Samet},
  journal={arXiv preprint arXiv:2411.08249},
  year={2024}
}

@article{ning2026ts,
  title={Ts-rag: Retrieval-augmented generation based time series foundation models are stronger zero-shot forecaster},
  author={Ning, Kanghui and Pan, Zijie and Liu, Yu and Jiang, Yushan and Zhang, James and Rasul, Kashif and Schneider, Anderson and Ma, Lintao and Nevmyvaka, Yuriy and Song, Dongjin},
  journal={Advances in Neural Information Processing Systems},
  volume={38},
  pages={163170--163199},
  year={2026}
}

@article{zhou2026stationarity,
  title={Stationarity-Aware Retrieval-Augmented Time Series Forecasting},
  author={Zhou, Shiqiao and Sch{\"o}ner, Holger and Wu, Zipeng and Fouch{\'e}, Edouard and Wilson, IAG and Wang, Shuo},
  journal={arXiv preprint arXiv:2606.04135},
  year={2026}
}

@article{yu2026memts,
  title={MEMTS: Internalizing Domain Knowledge via Parameterized Memory for Retrieval-Free Domain Adaptation of Time Series Foundation Models},
  author={Yu, Xiaoyun and Qiu, Xiangfei and Dong, Nanqing and Huang, Yonggui and Qi, Honggang and Pu, Geguang and Ouyang, Wanli and Chen, Xi and Hu, Jilin and others},
  journal={arXiv preprint arXiv:2602.13783},
  year={2026}
}

@misc{liu2026sundial,
      title={Sundial: A Family of Highly Capable Time Series Foundation Models}, 
      author={Yong Liu and Guo Qin and Zhiyuan Shi and Zhi Chen and Caiyin Yang and Xiangdong Huang and Jianmin Wang and Mingsheng Long},
      year={2026},
      eprint={2502.00816},
      archivePrefix={arXiv},
      primaryClass={cs.LG},
      url={https://arxiv.org/abs/2502.00816}, 
}

@misc{liu2026rationale,
      title={Rationale-Grounded In-Context Learning for Time Series Reasoning with Multimodal Large Language Models}, 
      author={Qingxiang Liu and Zhiqing Cui and Xiaoliang Luo and Yuqian Wu and Zhuoyang Jiang and Huaiyu Wan and Sheng Sun and Lvchun Wang and Wei Yu and Yuxuan Liang},
      year={2026},
      eprint={2601.02968},
      archivePrefix={arXiv},
      primaryClass={cs.AI},
      url={https://arxiv.org/abs/2601.02968}, 
}

@inproceedings{he2023domain,
  title={Domain adaptation for time series under feature and label shifts},
  author={He, Huan and Queen, Owen and Koker, Teddy and Cuevas, Consuelo and Tsiligkaridis, Theodoros and Zitnik, Marinka},
  booktitle={International conference on machine learning},
  pages={12746--12774},
  year={2023},
  organization={Pmlr}
}

@article{wu2022timesnet,
  title={Timesnet: Temporal 2d-variation modeling for general time series analysis},
  author={Wu, Haixu and Hu, Tengge and Liu, Yong and Zhou, Hang and Wang, Jianmin and Long, Mingsheng},
  journal={arXiv preprint arXiv:2210.02186},
  year={2022}
}

@article{wu2021autoformer,
  title={Autoformer: Decomposition transformers with auto-correlation for long-term series forecasting},
  author={Wu, Haixu and Xu, Jiehui and Wang, Jianmin and Long, Mingsheng},
  journal={Advances in neural information processing systems},
  volume={34},
  pages={22419--22430},
  year={2021}
}

@article{wu2025beyond,
  title={Beyond regularity: modeling chaotic mobility patterns for next location prediction},
  author={Wu, Yuqian and Peng, Yuhong and Yu, Jiapeng and Liu, Xiangyu and Yan, Zeting and Lin, Kang and Su, Weifeng and Qu, Bingqing and Lee, Raymond and Yang, Dingqi},
  journal={arXiv preprint arXiv:2509.11713},
  year={2025}
}

@inproceedings{zhou2021informer,
  title={Informer: Beyond efficient transformer for long sequence time-series forecasting},
  author={Zhou, Haoyi and Zhang, Shanghang and Peng, Jieqi and Zhang, Shuai and Li, Jianxin and Xiong, Hui and Zhang, Wancai},
  booktitle={Proceedings of the AAAI conference on artificial intelligence},
  volume={35},
  number={12},
  pages={11106--11115},
  year={2021}
}

@inproceedings{zhao2025proactive,
  title={Proactive model adaptation against concept drift for online time series forecasting},
  author={Zhao, Lifan and Shen, Yanyan},
  booktitle={Proceedings of the 31st ACM SIGKDD Conference on Knowledge Discovery and Data Mining V. 1},
  pages={2020--2031},
  year={2025}
}

@article{qiao2026multi,
  title={Multi-scale finetuning for encoder-based time series foundation models},
  author={Qiao, Zhongzheng and Liu, Chenghao and Zhang, Yiming and Jin, Ming and Pham, Quang and Wen, Qingsong and Suganthan, Ponnuthurai and Jiang, Xudong and Ramasamy, Savitha},
  journal={Advances in Neural Information Processing Systems},
  volume={38},
  pages={22313--22345},
  year={2026}
}

@inproceedings{ruan2025st,
  title={ST-LoRA: Low-rank adaptation for spatio-temporal forecasting},
  author={Ruan, Weilin and Chen, Wei and Dang, Xilin and Zhou, Jianxiang and Li, Weichuang and Liu, Xu and Liang, Yuxuan},
  booktitle={Joint European Conference on Machine Learning and Knowledge Discovery in Databases},
  pages={345--361},
  year={2025},
  organization={Springer}
}

@inproceedings{nie2022patchtst,
  title={A Time Series is Worth 64 Words: Long-term Forecasting with Transformers},
  author={Nie, Yuqi and Nguyen, Nam H. and Sinthong, Phanwadee and Kalagnanam, Jayant},
  booktitle={ICLR}, year={2023}}

@inproceedings{ma2026quizsf,
  title={QuiZSF: A Retrieval-Augmented Framework for Zero-Shot Time Series Forecasting},
  author={Ma, Shichao and Zhou, Zhengyang and Huang, Qihe and Wang, Binwu and Wang, Yang},
  booktitle={Proceedings of the ACM Web Conference 2026},
  pages={1947--1958},
  year={2026}
}

@inproceedings{jiang2026cross,
  title={Cross-city Time Series Forecasting with Retrieval-Augmented Large Language Models},
  author={Jiang, Yue and Liu, Chenxi and Chen, Yile and Chao, Qin and Liu, Shuai and Long, Cheng and Cong, Gao},
  booktitle={Proceedings of the ACM Web Conference 2026},
  pages={4962--4973},
  year={2026}
}

@inproceedings{fan2026automated,
  title={Automated Model Selection for Multivariate Time Series Forecasting},
  author={Fan, Xiaoxuan and Sun, Jiaqi and Deng, Xianjun and Zhang, Qiankun and Xiang, Wei and Liu, Shenghao and Yi, Lingzhi},
  booktitle={Proceedings of the ACM Web Conference 2026},
  pages={5322--5333},
  year={2026}
}

@inproceedings{zhang2026semixer,
  title={Semixer: Semantics enhanced mlp-mixer for multiscale mixing and long-term time series forecasting},
  author={Zhang, Xu and Wang, Qitong and Wang, Peng and Wang, Wei},
  booktitle={Proceedings of the ACM Web Conference 2026},
  pages={5636--5647},
  year={2026}
}

@article{deng2025lm,
  title={LM-mixup: Text Data Augmentation via Language Model based Mixup},
  author={Deng, Zhijie and Shen, Zhouan and Li, Ling and Zhou, Yao and Zhu, Zhaowei and He, Yanji and Wang, Wei and Wei, Jiaheng},
  journal={arXiv preprint arXiv:2510.20449},
  year={2025}
}

@article{deng2026codeblock,
  title={CODEBLOCK: Learning to Supervise Code at the Right Granularity},
  author={Deng, Zhijie and Li, Ling and Pang, Jinlong and Hu, Kaiqin and Xuan, Qi and Zhu, Zhaowei and Wei, Jiaheng},
  journal={arXiv preprint arXiv:2606.18286},
  year={2026}
}

@misc{wu2026lifeside,
      title={LifeSide: Benchmarking Agents as Lifelong Digital Companions}, 
      author={Yuqian Wu and Zhijie Deng and Wei Chen and Junwei Li and Yutian Jiang and Junle Chen and Zhengjun Huang and Qingxiang Liu and Jing Tang and Jiaheng Wei and Yuxuan Liang},
      journal={arXiv preprint arXiv:2606.04660},
      year={2026},
}

@article{wu2026back,
  title={Back to basics: Let conversational agents remember with just retrieval and generation},
  author={Wu, Yuqian and Chen, Wei and Huang, Zhengjun and Chen, Junle and Liu, Qingxiang and Wang, Kai and Zhou, Xiaofang and Liang, Yuxuan},
  journal={arXiv preprint arXiv:2604.11628},
  year={2026}
}
